%% file: main.tex
\documentclass[lettersize,journal]{IEEEtran}
\usepackage{amsmath,amsfonts}
\usepackage{algorithmic}
\usepackage{array}
\usepackage[caption=false,font=normalsize,labelfont=sf,textfont=sf]{subfig}
\usepackage{textcomp}
\usepackage{stfloats}
\usepackage{url}
\usepackage{verbatim}
\usepackage{cite}
\usepackage{pifont}
\usepackage{bbm}
\usepackage{amsmath}
\usepackage{amssymb}
\usepackage{booktabs}
\usepackage{multirow}
\usepackage{xcolor}
\usepackage{url}
\usepackage{caption}
\usepackage{subcaption}

\usepackage{colortbl}
\definecolor{mygray}{rgb}{1,0.95,0.97}

\definecolor{LightCyan}{RGB}{0.88,1,1}
\definecolor{sgreen}{RGB}{30, 150, 30} 
\definecolor{cvprblue}{RGB}{54,125,189}

\usepackage[pagebackref,breaklinks,colorlinks,allcolors=blue]{hyperref}

\usepackage[capitalize]{cleveref}
\crefname{section}{Sec.}{Secs.}
\Crefname{section}{Section}{Sections}
\Crefname{table}{Table}{Tables}
\crefname{table}{Tab.}{Tabs.}

\usepackage{bm}
\usepackage{amsthm,amsmath,amssymb}
\usepackage{mathrsfs}
\usepackage{booktabs}
\usepackage{multicol}
\usepackage{float}
\usepackage{subfig}
\usepackage{enumitem}
\usepackage{bbding}
\usepackage{mathrsfs}
\usepackage{color}
\usepackage{colortbl}
\usepackage{rotating}
\usepackage{array}

\definecolor{LightCyan}{rgb}{0.88,1,1}

\usepackage{adjustbox}

\usepackage{tabularx}
\usepackage{etoolbox}
\usepackage{makecell}
\usepackage{pifont}

\usepackage{wrapfig}
\usepackage{cutwin}
\usepackage{alltt}
\usepackage{graphicx}
\usepackage{enumitem}
\usepackage{bbding}

\usepackage{mathrsfs}
\usepackage{booktabs}
\usepackage{multicol}
\usepackage{color}
\usepackage{float}
\usepackage{subfig}
\usepackage{colortbl}
\usepackage{rotating}
\usepackage{array}

\definecolor{LightCyan}{rgb}{0.88,1,1}

\usepackage[ruled]{algorithm2e}

\SetAlgoNlRelativeSize{0}

\begin{document}

\title{eMotions: A Large-Scale Dataset and Audio-Visual Fusion Network for Emotion Analysis in Short-form Videos}

\author{Xuecheng Wu, Dingkang Yang, Danlei Huang, Xinyi Yin, Yifan Wang, Jia Zhang, Jiayu Nie \\ Liangyu Fu, Yang Liu, Junxiao Xue, Hadi Amirpour, and Wei Zhou
\thanks{This research was supported by the Zhejiang Provincial Natural Science Foundation of China under Grant No. LQ23F030009.}
\thanks{Xuecheng Wu, Danlei Huang, and Jia Zhang are with the School of Computer Science and Technology, Xi'an Jiaotong University, Xi'an, 710049, China. (E-mail: {\small wuxc3@stu.xjtu.edu.cn});}
\thanks{Dingkang Yang is with the College of Intelligent Robotics and Advanced Manufacturing, Fudan University~\&~ByteDance, Shanghai, 200433, China (E-mail: {\small yangdingkang@bytedance.com});}
\thanks{Xinyi Yin is with the School of Cyber Science and Engineering, Zhengzhou University, Zhengzhou, 450002, China (E-mail: {\small yinxinyi@stu.zzu.edu.cn});}
\thanks{Yifan Wang is with the Institute of Advanced Technology, University of Science and Technology of China, Hefei, 230031, China (E-mail: {\small wangyfan@mail.ustc.edu.cn});}
\thanks{Jiayu Nie is with the Inspur Electronic Information Industry Co., Ltd, Jinan, 250101, China. (E-mail: {\small niejiayu@inspur.com}).}
\thanks{Liangyu Fu is with the School of Software, Northwestern Polytechnical University, Xi'an, 710072, China. (E-mail: {\small lyfu@mail.nwpu.edu.cn}).}
\thanks{Yang Liu is with the Department of Computer Science, The University of Toronto, Toronto, ONM5S1A1, Canada. (E-mail: {\small yangliu@cs.toronto.edu});}
\thanks{Junxiao Xue is with the Research Center for Space Computing System, Zhejiang Lab, Hangzhou, 311100, China (E-mail: {\small xuejx@zhejianglab.cn});}
\thanks{Hadi Amirpour is with the Institute of Information Technology, University of Klagenfurt, Klagenfurt, 9020, Austria. (E-mail: {\small Hadi.Amirpour@aau.at});}
\thanks{Wei Zhou is with the School of Computer Science and Informatics, Cardiff University, Cardiff, CF24 4AG, United Kingdom. (E-mail: {\small zhouw26@cardiff.ac.uk});}
\thanks{Xuecheng Wu \& Dingkang Yang deserve equal contributions.}
\thanks{Corresponding authors: Junxiao Xue \& Yang Liu.}
\thanks{Manuscript received XX, 2025; revised XX, 2025.}
}

\markboth{IEEE Transactions on Circuits and Systems for Video Technology}%
{Shell \MakeLowercase{\textit{et al.}}: A Sample Article Using IEEEtran.cls for IEEE Journals}

\maketitle

\begin{figure*}[t!]
\setlength{\belowcaptionskip}{-0.1cm}
\includegraphics[width=\textwidth]{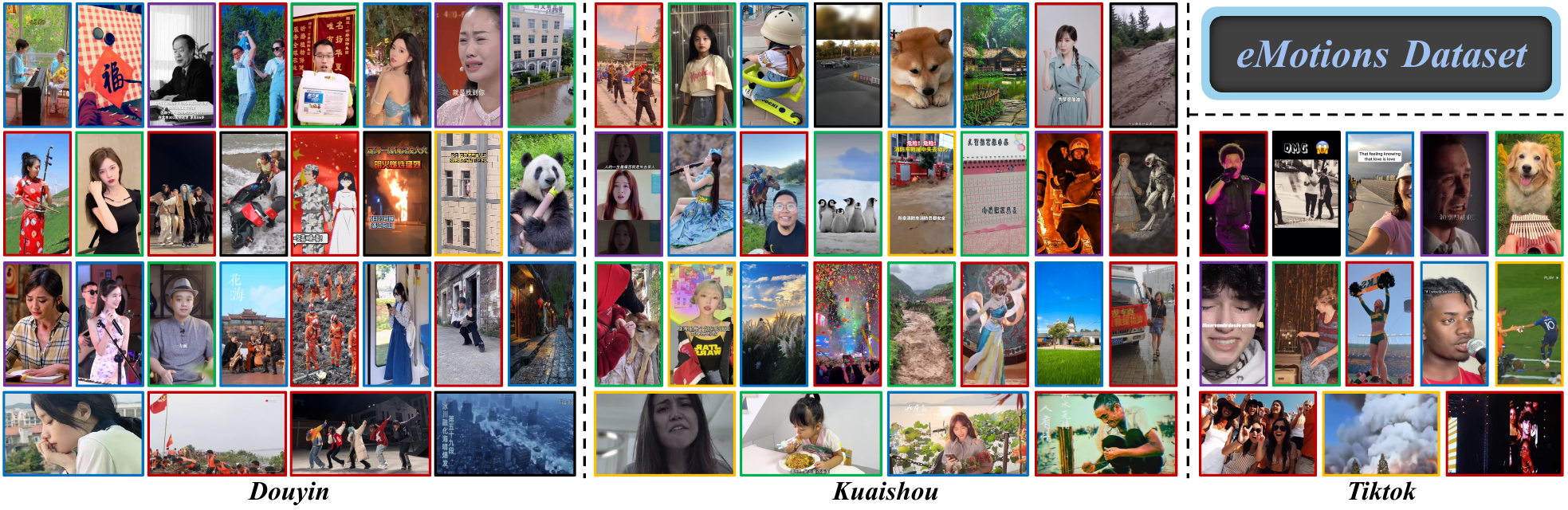}
\vspace{-1.4em}
\captionof{figure}{An overview of eMotions composed of 27,996 videos of six emotions across Douyin, Kuaishou, and Tiktok. The colors of frame borders specify the categories to which they belong: \textit{\textcolor[RGB]{192,0,0}{Excitation}}, \textit{\textcolor[RGB]{0,0,0}{Fear}}, \textit{\textcolor[RGB]{0,176,80}{Neutral}}, \textit{\textcolor[RGB]{0,112,192}{Relaxation}}, \textit{\textcolor[RGB]{112,48,160}{Sadness}}, \textit{\textcolor[RGB]{255,192,0}{Tension}}.}
\vspace{-0.5em}
\label{fig1}
\end{figure*}

\input{Sec/0_abs}


\input{Sec/1_intro}

\input{Sec/2_related}
\input{Sec/3_dataset}

\input{Sec/3d5_method}
\input{Sec/4_expers}

\input{Sec/5_conclusions}

\bibliographystyle{IEEEtran}
\bibliography{main}

\end{document}

%% file: Sec/0_abs.tex
\begin{abstract}
Short-form videos (SVs) have become a vital part of our online routine for acquiring and sharing information. Their multimodal complexity poses new challenges for video analysis, highlighting the need for video emotion analysis (VEA) within the community. Given the limited availability of SVs emotion data, we introduce eMotions, a large-scale dataset consisting of 27,996 videos with full-scale annotations. To ensure quality and reduce subjective bias, we emphasize better personnel allocation and propose a multi-stage annotation procedure. Additionally, we provide the category-balanced and test-oriented variants through targeted sampling to meet diverse needs. While there have been significant studies on videos with clear emotional cues (\textit{e.g.}, facial expressions), analyzing emotions in SVs remains a challenging task. The challenge arises from the broader content diversity, which introduces more distinct semantic gaps and complicates the representations learning of emotion-related features. Furthermore, the prevalence of audio-visual co-expressions in SVs leads to the local biases and collective information gaps caused by the inconsistencies in emotional expressions. To tackle this, we propose AV-CANet, an end-to-end audio-visual fusion network that leverages video transformer to capture semantically relevant representations. We further introduce the Local-Global Fusion Module designed to progressively capture the correlations of audio-visual features. Besides, EP-CE Loss is constructed to globally steer optimizations with tripolar penalties. Extensive experiments across three eMotions-related datasets and four public VEA datasets demonstrate the effectiveness of our proposed AV-CANet, while providing broad insights for future research. Moreover, we conduct ablation studies to examine the critical components of our method. Dataset and code will be made available at \href{https://github.com/XuecWu/eMotions}{Github}.
\end{abstract}

%% file: Sec/1_intro.tex
\section{Introduction}
\label{sec:introduction}

\IEEEPARstart{V}{ideo} emotion analysis (VEA) focuses on interpreting the meaning of various elements and determining the emotions they evoke in viewers. Meanwhile, the emotions of viewers can be influenced by various elements, such as videos, audio, and text from the online streaming media \cite{new_21_liu2022ser30k,yang2022disentangled}, which has led to the advancements of multi-modal VEA. In particular, short-form videos (\href{https://en.wikipedia.org/wiki/Video_clip\#Short-form_videos}{SVs}), a new form of social media tool, have made rapid progress in recent years. SVs are succinct and expressive, combining visual, auditory, and other elements to intensify emotional expressions and elicit emotional resonance from viewers, which is crucial to spreading emotions. Thus, conducting VAE towards SVs—serving as a crucial task in video understanding—has become an important research direction in video processing systems, with remarkable application value in diverse fields such as opinion mining \cite{new_54_sobkowicz2012opinion}, dialogue systems~\cite{5_schuller2018age} and healthcare~\cite{depression_recognition}.

Although VEA in facial expressions \cite{new_1_wang2020suppressing,new_3_yang2021circular,facial_expression_recognition} has been extensively studied, the research in SVs remains light due to the absence of large-scale datasets. To address this gap and facilitate further research, we introduce eMotions (Fig.~\ref{fig1}), a dataset specifically designed for emotion analysis in short-form videos, which is the first large-scale dataset with full-scale annotations in this field. eMotions consists of 27,996 videos with corresponding audio from Douyin, Kuaishou, and Tiktok three SVs platforms, covering various contents across diverse dimensions and totaling almost 198 hours of total footage. Specifically, considering the discrete emotional categories in psychology and the content distribution characteristics of eMotions, we label each sample using the six emotional categories proposed by Plutchik in \cite{1_plutchik1994psychology} (\textit{i.e.}, Excitation, Fear, Neutral, Relaxation, Sadness, Tension). Moreover, to alleviate the impact of subjectivities on annotation quality, we elaborately adjust the personnel allocations through proposed two-stage Cross-Check and consistencies evaluations, as well as introduce a multi-stage annotation workflow. In addition, catering to the class distribution of eMotions and the testing demands of research community, we provide the category-balanced and test-oriented variant datasets.

\begin{figure}[t!]
\setlength{\belowcaptionskip}{-0.5cm}
\centering
\includegraphics[width=\linewidth]{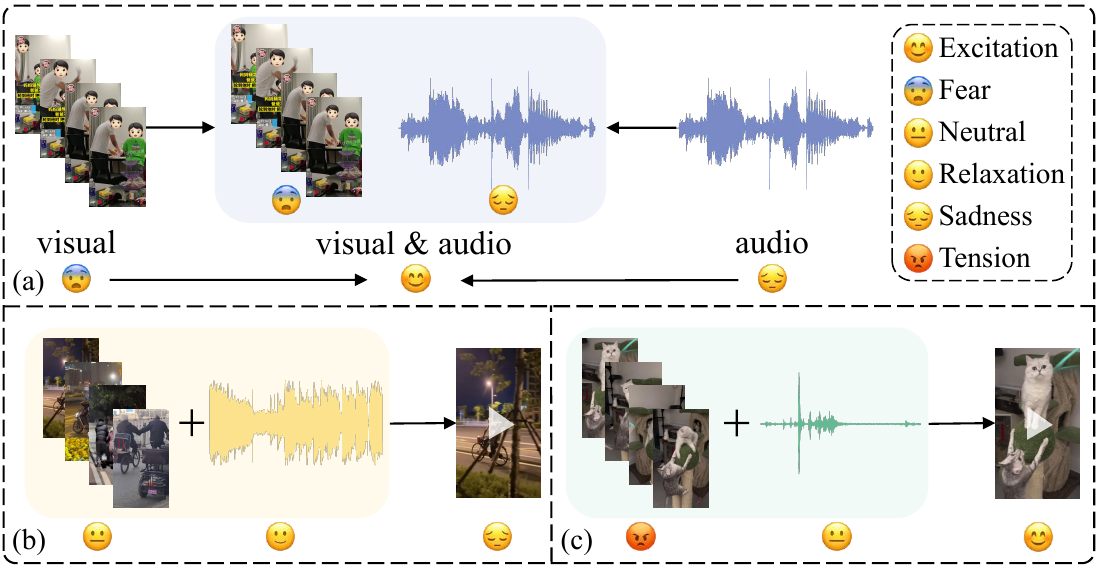}
\caption{The illustrations of emotion inconsistence. \textbf{(a)} Separate visual or auditory modality evokes conflicting emotions. \textbf{(b)}~\&~\textbf{(c)} The lack of emotional information evoked from visual or auditory modality results in emotion disalignment.}
\vspace{-0.15em}
\label{fig_introd_challenge}
\end{figure}

Different from traditional video emotion analysis (VEA), VEA towards SVs features the following challenges: As displayed in Fig.~\ref{fig1}, the content diversity in SVs is broader, which leads to more distinct semantic gaps and barriers of learning emotion-related features than commonly used videos. At the same time, there exists emotion inconsistence under the prevalent audio-visual co-expressions, as shown in Fig.~\ref{fig_introd_challenge}. As a result, at the local-level, inconsistence of emotion evocations with difficulties in integrating information lead to biases in representations learning. Additionally, at the global-level, the obstacles in emotion understanding within the overall feature space and accumulation of local biases result in more collectively significant information gaps. Considering these observations, we analyze three design keypoints in the following: \textbf{(1)} Learning more semantically relevant representations and emotion-related features in an end-to-end manner. \textbf{(2)} Interactive correlated modeling of inter-modalities at local-level, followed by selective integrations. \textbf{(3)} Supplementally facilitating complementary modeling of audio-visual correlations at global-level, while learning more contextual representations.

Based on the above design keypoints, we present an end-to-end audio-visual network denoted AV-CANet (\textbf{A}udio-\textbf{V}iusal \textbf{C}ooperatively enhanced \textbf{A}nalysis \textbf{Net}work) as the baseline method on the proposed eMotions dataset. Unlike previous CNNs-based VEA methods \cite{35_zhao2020end,tcsvt_CNN-based}, we employ Video Swin-Transformer \cite{19_liu2022video} as the visual backbone, which makes AV-CANet naturally capture the global relations between regions in each frame and efficiently model the long-range dependencies as well as long-term sequences, leading to more semantically relevant representations. The Local-Global Fusion Module (LGF Module) is designed to mitigate the local biases and collective information gaps, which progressively captures the correlated information of inter-modalities to output more comprehensive representations, as displayed in Fig.~\ref{fig5-AV-CANet} below. Besides, we propose the EP-CE Loss (\textbf{E}motion \textbf{P}olarity enhanced \textbf{C}ross-\textbf{E}ntropy Loss), which incorporates three emotion polarities (\textit{i.e.}, positive, neutral, negative) to guide model focusing more emotion-related features. The proposed method achieves superior performance in comparisons with recently advanced VEA baselines across three eMotions-related and four public datasets, indicating that it has the capability to benchmark eMotions. In summary, the proposed eMotions dataset and AV-CANet can provide foundations and broad insights for the development of future works in emotion analysis towards short-form videos. Additionally, beyond emotion analysis, the abundant emotions conveyed in eMotions can facilitate research in LLMs and MLLMs (\textit{e.g.}, synchronous alignments with human behaviors~\cite{ViC-Bench}), emotional content generation~\cite{new_55_yang2024emogen,AVF-MAE++,yang2024asynchronous}, and explainable emotional reasoning~\cite{new_56_lian2023explainable}. Our AV-CANet, as a competitive audio-visual joint fusion network, can also offer new insights into the development of general audio-visual scenarios modeling.

This paper significantly extends from our preliminary conference version~\cite{wu2025towards}. We provide multi-faceted enhancements to further strengthen our work and promote more far-reaching impacts. To be specific, \textbf{(1)} We introduce the newly proposed personnel adjustment strategy for our dataset construction procedure, along with the pseudo code and overall two-stage cross-check experimental results in supplementary material to validate the effectiveness of our adjustments. We hope our strategy can offer a new feasible solution for improving the emotional data annotation quality in future research; \textbf{(2)} In this version, we present the significantly enhanced discussions of our motivation, dataset, network architecture, experimental results, as well as the implementations; \textbf{(3)} We perform more comprehensive qualitative and quantitative experiments on the proposed audio-visual local-global fusion module to investigate the effectiveness and robustness of our design against current advanced audio-visual fusion methods; \textbf{(4)} We provide extensive qualitative and quantitative experimental results and analysis at both the dataset and method levels, aiming to lay foundations for the future developments of emotion analysis methods towards short-form videos (SVs); \textbf{(5)} Starting from the learning anchor point, we thoroughly explore the differences between VEA towards SVs and that for other application scenarios, providing a deeper understanding of the necessity of emotion analysis for SVs; \textbf{(6)} Considering the rapid progress of multimodal large language models (MLLMs), we investigate the possibility of utilizing MLLMs for short-form video emotion analysis and conclude the current limitations and potentials for further developments.

%% file: Sec/2_related.tex
\section{Related Works}
\label{sec:related}

\subsection{Video Emotion Analysis Datasets} 
As video emotion analysis (VEA) progresses, a growing number of datasets have been introduced to support research in this emerged research field. FABO \cite{new_9_gunes2006bimodal} comprises 1.9\textit{k} video clips capturing both facial and body expressions, recorded through cameras. IEMOCAP \cite{26_busso2008iemocap}, one of the earliest audio-visual datasets for VEA, was filmed in the laboratory setting. AFEW \cite{new_6_2012Collecting} includes 1,426 videos from 330 participants, each annotated with one of seven emotions. Aff-Wild2 \cite{new_8_kollias2018aff} features 558 videos annotated for valence-arousal estimations. VideoEmotion8 \cite{12_jiang2014predicting} and Ekman6 \cite{13_xu2016video} contain 1,101 and 1,637 clips, respectively, sourced from online video platforms. CMU-MOSEI \cite{16_zadeh2018multimodal} consists of 23,500 utterances from YouTube, categorized into six emotional classes. CAER \cite{18_lee2019context} and MELD \cite{33_soujanya2018multimodal} consist of 13,201 and 13,708 clips, respectively, derived from TV shows. Music\_video \cite{7_pandeya2021deep} holds 3,323 music videos labeled with six emotional categories. DFEW \cite{new_58_jiang2020dfew} contains 16,372 clips extracted from thousands of movies, spanning across seven emotional categories. MAFW \cite{new_57_liu2022mafw} is a comprehensive facial emotion database, comprising 10,045 video clips. MER2024 \cite{new_42_lian2024mer} includes 5,030 labeled video clips and 115\textit{k} unlabeled clips, spanning six emotional categories and organized into three subsets. Compared to previously published datasets, eMotions demonstrates the following notable features: \textbf{(1)} It represents the first large-scale video emotion analysis (VEA) dataset tailored for short-form videos (SVs), and remains the largest VEA dataset with full-scale video-level annotations. \textbf{(2)} An emphasis on better personnel allocations and multi-stage annotations to reduce the influence of subjectivities on emotional annotation quality, engaging 12 annotators and one expert. \textbf{(3)} The larger extent of content diversity, covering a multi-cultural spectrum and spanning an extensive timeline of 45 months. \textbf{(4)} Additionally, two tailored variants of our eMotions dataset are established through targeted sampling to support diverse research needs.

\begin{figure*}[t!]
\setlength{\belowcaptionskip}{-1.5em}
\centering
\includegraphics[width=\textwidth]{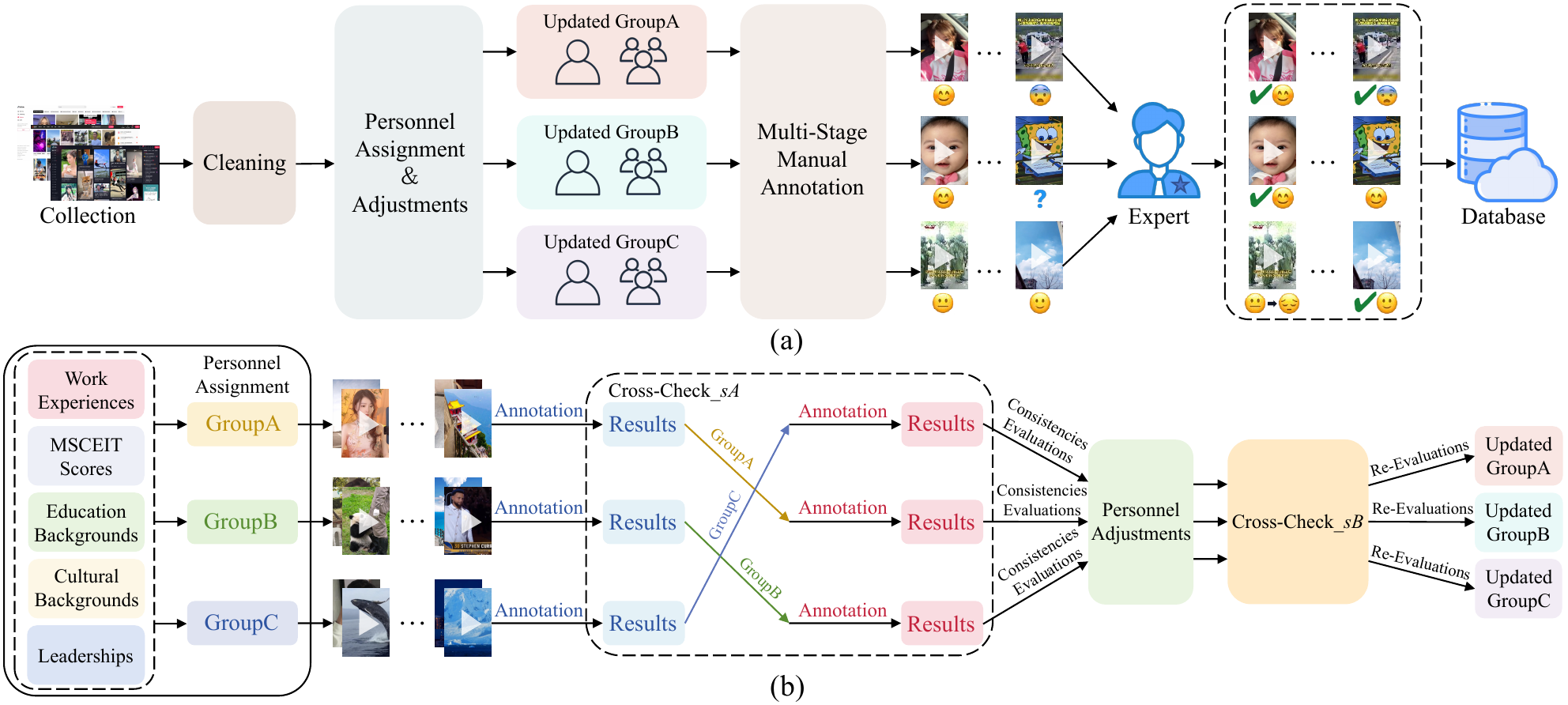}
\vspace{-1.0em}
\caption{\textbf{(a)} The overall pipeline of dataset construction. \textbf{(b)} The detailed workflow of personnel assignment and adjustments.} 
\label{fig_pipeline_adjustment}
\end{figure*}

\subsection{Video Emotion Analysis Methods} 
Early research on video emotion analysis (VEA) primarily focused on the handcrafted feature design 
\cite{12_jiang2014predicting}. The growing availability and diversity of VEA datasets have substantially accelerated the advancement of deep learning approaches in video emotion analysis. For uni-modal approaches, Zhao et al. \cite{new_34_zhao2021former} proposes a dual-branch framework incorporating CS-Former and T-Former modules. \cite{new_12_xue2022coarse} developes a coarse-to-fine cascaded network with smooth prediction capability. Zhao et al. \cite{new_36_zhao2023prompting} enhances model performance by integrating CLIP \cite{new_43_radford2021learning} and learnable prompting tokens. In the domain of audio-visual emotion analysis, \cite{40_zhang2023weakly} introduces a temporal erasing network that can facilitate context-aware learning by focusing on salient keyframes in audio-visual sequences. \cite{35_zhao2020end} employs a combination of attention mechanisms to improve the quality of audio-visual feature representations. \cite{new_13_mocanu2023multimodal} introduces an intra-modal attention network to produce more fine-grained features. \cite{new_35_su2020msaf} proposes a lightweight cross-modal fusion module designed to emphasize features with higher emotional relevance. To model the interplay between facial and auditory behaviors, Tran et al. \cite{new_16_tran2022pre} develops a pre-training framework tailored for audio-visual data. \cite{new_14_praveen2023audio} presents a unified model that can extract emotionally salient features from both audio and visual modalities. Peng et al. \cite{50_peng2022balanced} and Xu et al. \cite{53_xu2023mmcosine} enhance performance through adaptive gradient modulation and a novel cosine-based loss function, respectively. To address modality incompleteness, \cite{new_18_chumachenko2022self} designs a new model capable of handling missing audio or visual data. To capture context-aware semantics, Poria et al. \cite{55_poria2017context} introduces a model grounded in multiple LSTM units. \cite{new_59_zhang2024mart} develops a MAE-style VEA method via token masking. \cite{new_17_schoneveld2021leveraging} leverages the teacher-student networks to extract visual features, which are subsequently integrated with the audio representations through model-level fusion. In this paper, we propose AV-CANet, a novel network specifically designed to address the inherent challenges of our eMotions dataset. Extensive experiments demonstrate that our model can achieve superior performance across three eMotions-related subsets and four public VEA datasets.

%% file: Sec/3_dataset.tex
\section{Dataset Construction}
\label{sec:emotions}

The overall construction pipeline of eMotions is illustrated as Fig.~\ref{fig_pipeline_adjustment}~(a), involving data collection and cleaning (Sec.~\ref{subsec:3.1}), personnel assignment and adjustments (Sec.~\ref{subsec:3.2} \& \ref{PAS}), multi-stage manual annotation and expert re-review (Sec.~\ref{subsec:3.3}). We then present the labeling quality evaluations and dataset characteristics (Sec.~\ref{subsec:3.4}~\&~\ref{subsec:3.5}). Furthermore, we provide the construction details for two variants of our introduced eMotions dataset (Sec.~\ref{subsec:3.6}).

\subsection{Data Collection and Cleaning}
\label{subsec:3.1}

To initiate the data acquisition process, we begin by collecting trending topics and identifying target user groups from short-form videos (SVs) platforms, utilizing Selenium to render dynamically generated content. Full-page URLs for each video are then extracted using Xpath. Following User-Agent spoofing, each URL is accessed to retrieve detailed meta-data associated with the video. Subsequently, regular expressions and decoding functions are applied to extract video titles and corresponding download links. The videos are then downloaded, again employing user-agent spoofing, and stored in binary format at designated locations. Finally, videos that do not meet the predefined duration criteria are removed, and the remaining videos are systematically renamed in numerical order. Furthermore, the extracted meta-data is leveraged to mine salient keywords and update the repository of trending terms, which in turn guides iterative refinement of the overall crawling pipeline. Besides, a staged crawling strategy is adopted, running in parallel with the annotation workflow to ensure temporal diversity and coverage across the dataset. During the data cleaning phase, duplicate and corrupted files are removed as an initial step. Videos containing content related to racial discrimination, violence, or NSFW are filtered out through a combination of automated detection and manual review, in order to mitigate potential ethical biases in the model training. In addition, to preserve the integrity of discrete emotional labeling, videos that exhibit continuous or ambiguous emotional transitions are excluded from our final constructed dataset.

\subsection{Personnel Assignment and Adjustments}
\label{subsec:3.2}

In our procedure, each annotator is asked first to pass the labeling test, comprising the sentiment quotient test and the annotation quality evaluation \cite{new_21_liu2022ser30k}. When scoring 90 or above in accuracy, the annotators can take part in the emotional annotation. We then hold training sessions for them, including the detailed introduction of annotation workflow and targeted learning on multi-cultures.

\noindent \textbf{Assignment:} We empirically consider five important factors and set their coefficients to determine the assignment of group members and leaders. We also balance the gender distribution by allocating two males and one female for each annotation group. The overall process can be formulated as:
\begin{equation}
p = \alpha_\textit{1} \cdot we + \alpha_\textit{2} \cdot ms + \alpha_\textit{3} \cdot (eb + cb + lp),
\label{eq1}
\end{equation}
where $we$, $eb$, $cb$, and $lp$ respectively refer to the work experiences, education backgrounds, cultural backgrounds, and leaderships, which are all quantified using scoring tests. For MSCEIT \cite{2_mayer2002mayer} scores ($ms$), higher ranking indicates better emotional cognition. $\alpha_{i}$ ($i \in \{1, 2, 3 \}$) = $\{0.4, 0.3, 0.1\}$ are the weight coefficients, respectively.

\begin{table}[t!]
\centering
\renewcommand{\arraystretch}{1.2}
\setlength{\arrayrulewidth}{0.22pt}
\vspace{0.3em}
\caption{The evaluation results of consistencies of intra-group and inter-group ($S_a$ and $S_r$) following Cross-Check.} 
\resizebox{\linewidth}{!}{%
{\tiny
\begin{tabular}{@{}ccccccc@{}}
\toprule[0.22pt]
\hspace{0.08cm}\multirow{2}{*}{\raisebox{-1ex}{Stage}} & \multicolumn{2}{c}{GroupA} & \multicolumn{2}{c}{GroupB} & \multicolumn{2}{c}{GroupC} \\ \cmidrule[0.22pt](l){2-7} 
                                & $S_a$  & $S_r$ & $S_a$  & $S_r$ & $S_a$  & $S_r$ \\ \midrule[0.22pt]
\textit{sA}                       & 0.52        & 52.55       & 0.53        & 48.78       & 0.55        & 54.54       \\ 
\textit{sB}                        & 0.55        & 55.53       & 0.56        & 54.41       & 0.56        & 55.79       \\ \bottomrule[0.22pt]
\end{tabular}
}
}
\vspace{-1.3em}
\label{tab5}
\end{table}

\noindent \textbf{Adjustments:} We perform the personnel adjustments following assignment to alleviate the impact of subjectivities on the annotation workflow employing multi-groups with multi-annotators, which are reflected in the improvements of the consistencies of intra-group and inter-group (\textit{i.e.}, $S_a$, $S_r$) \cite{new_29_lavitas2021annotation}. Specifically, we select 9,000 samples from the cleaned data and distribute them evenly among the assigned groups. GroupA (\textit{GA}), GroupB (\textit{GB}), and GroupC (\textit{GC}) are each requested to perform annotation following the workflow described in Sec.~\ref{subsec:3.3} below, and the leaders here only carry out result collections. We then sample 18 sets from the annotations of each group to conduct two-stage Cross-Check \cite{new_28_abercrombie2023consistency,new_27_hallgren2012computing}, in which we first exchange the annotations of three groups in pairs and then label these samples again, as illustrated in Fig.~\ref{fig_pipeline_adjustment}~(b). These sets, each including 100 samples, are equally divided into two parts for two-stage Cross-Check (\textit{i.e.}, \textit{sA}, \textit{sB}). Each stage has three sets for Neutral, two sets for Excitation, and one set for each of Sadness, Relaxation, Tension, as well as Fear. Afterwards, we evaluate the consistencies of intra-group and inter-group under the present allocations based on the results of Cross-Check\textit{\_sA}. The ranges of $S_a$ and $S_r$ formulated as Eq.~\ref{eq2} and Eq.~\ref{eq4} below are from 0 to 1 and 0 to 70, respectively. The larger $S_a$ and $S_r$ indicate the increasing consistencies for annotations.
\begin{align}  
S_a &= \frac{1}{n} \cdot \sum_{i=0}^{n-1} \left(\frac{1}{3 \cdot m} \cdot \sum_{j=1}^{m}c_j\right),
\label{eq2} \\
S_r &= \frac{1}{c} \sum_{i=0}^{n-1} w_i \cdot\left(0.7 \cdot C_i+0.3 \cdot\left(m-{C_i}-{M_i}\right)\right), 
\label{eq4}
\end{align}
where $n$, $m$, and $c$ denote the number of sets, samples per set, and emotional categories, respectively. $c_j$ refers to the quantity of currently annotated categories of three annotators for one sample that are consistent with the previous emotional category. $w_i \in \{\frac{1}{3} \mid 0 \leq i<3, \frac{1}{2} \mid 3 \leq i <5, 1 \mid 5 \leq i \leq 8\}$ represents the weight coefficient for each set, depending on the number of sets in each category. $C_i$ stands for the samples consistent with previous annotations. $M_i$ refers to the ``more" samples, indicating the final annotation is indeterminate.

As shown in Tab.~\ref{tab5}, $S_a$ of the three groups achieve 0.52, 0.53, and 0.55, respectively. $S_r$ stand at 52.55, 48.78, and 54.54, with \textit{GB} scoring relatively low. Considering these observations, we perform targeted adjustments (as detailed in Sec.~\ref{PAS} below), then re-evaluate the consistencies after Cross-Check\textit{\_sB}. Following personnel adjustments, $S_a$ of \textit{GA} and \textit{GB} both rise by 0.03, and that of \textit{GC} improves by 0.01. $S_r$ of three groups increase by 2.98, 5.63, and 1.25, in which \textit{GB} exhibits the highest improvement.

\subsection{Personnel Adjustments Strategy}
\label{PAS}

To mitigate the impact of human subjectivity on emotional annotation quality, starting from the perspective of allocations, we newly design a personnel adjustments strategy (as can be seen in Algorithm~\textcolor{blue}{1} of the supplementary material) that can dynamically adjust the annotator allocation based on our two-stage cross-check outcomes. Specifically, our strategy first computes the consistency ratios ($CRs$) for each annotator in different categories. Annotators with the highest and lowest $CRs$ in each category are then swapped across groups to improve inter-group agreement. After the personnel adjustments for each category, the swaped annotators will re-annotate the overall sets from Cross-Check\textit{\_sA} in the updated groups. The procedure iterates over the six emotional categories and terminates when the average consistency improves and the inter-group variance falls below the threshold. After two weeks of adjustments, the mean $\mu$ increases from 51.96 to 55.24 and the variance $\sigma^2$ decreases from 5.72 to 0.36, indicating that the personnel allocations for three groups have become more balancedly consistent in emotional cognition. These results impressively indicates that we have finalized the personnel allocations that can be deployed for our formal annotation.

\subsection{Multi-Stage Manual Annotations}
\label{subsec:3.3}

Defining emotion categories is a critical step in constructing high-quality VEA datasets. We adopt the six basic emotions proposed by Plutchik \cite{1_plutchik1994psychology}, as they can offer clearer boundaries and effectively reflect the emotional dynamics described in the Russell's Circumplex Model \cite{new_60_posner2005circumplex}. The multi-stage manual annotation workflow combines member votes and leader evaluations, benefiting to alleviate the impact of subjectivities. Following \cite{7_pandeya2021deep}, we adopt and extend the mapping table of emotion category-to-adjective to promote annotations, as displayed in Tab.~\textcolor{blue}{2} of supplementary material. Benefit from this mapping table, the annotators only need to select the adjective that best match their emotional feelings. Moreover, to boost the stimulation and engagement of annotators, we develop a labeling interface, where one sample, annotation instructions, confidence evaluations, and emotional adjective selections are provided to assist annotators. Besides, this interface can automatically translate 62 adjective choices into six emotional categories. To be specific, the leader distributes data, followed by three members undertaking annotations via the proposed mapping table. Meanwhile, members are asked to attach confidence scores to their annotations, and the average confidence score finally reaches 0.7. Afterwards, the leader collects these annotations, leveraging a majority voting scheme to determine labels. If annotations from three members are all different (\textit{i.e.}, samples labeled ``more"), the leader will intervene in labeling. If a decisive majority of four votes emerges, the final labels can be directly determined. If consensus is still unreachable, leaders will exchange samples to facilitate decision-making. In five votes, a clear majority allows us to determine the final labels. If consensus continues to be inaccessible, the annotation expert from \href{https://cloud.baidu.com/product/dcs.html}{BD Cloud} will finalize the labels in re-review. After completing the overall annotation process, we calculate the overall Fleiss'kappa score, achieving $k >$ \textit{0.45}.

\begin{figure}[t!]
\setlength{\belowcaptionskip}{-0.5cm}
\centering
\includegraphics[width=\linewidth]{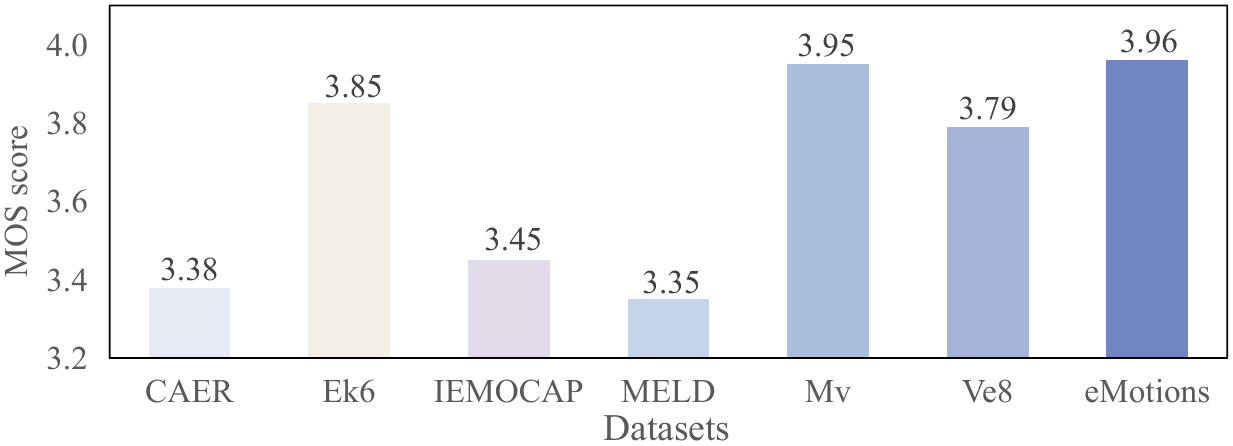} 
\vspace{-1.0em}
\caption{The MOS scores of six VEA datasets and our eMotions. Ek6: Ekman6. Mv: Music\_video. Ve8: VideoEmotion8.}
\label{fig_MOS_Scores}
\end{figure}

\begin{figure*}[t!]
\setlength{\belowcaptionskip}{-0.1cm}
\centering
\includegraphics[width=\textwidth]{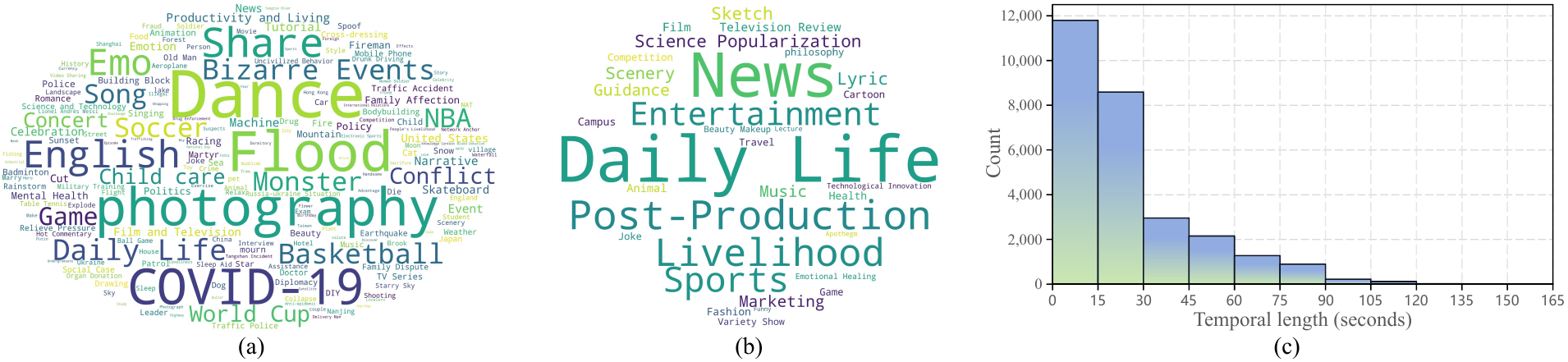}
\vspace{-1.2em}
\caption{\textbf{(a)} \& \textbf{(b)} Word clouds of topics and content types in our eMotions dataset. Larger text size indicates a higher frequency of occurrence. \textbf{(c)} Duration distribution of short-form videos in our dataset.}
\label{fig_wordclouds_duration}
\end{figure*}

\subsection{Labeling Quality Evaluations}
\label{subsec:3.4}

In this section, we evaluate the annotation quality of eMotions in comparison with six existing VEA datasets. A total of 979 samples are randomly selected across the seven datasets following the sample correction formula proposed by \cite{new_23_singh2014sampling}. Subsequently, four emotional annotation experts are invited to assess the labeling quality of these samples. We adopt MOS \cite{new_24_ribeiro2011crowdsourcing} as the evaluation metric, where each video is rated on a 5-point scale (\textit{i.e.}, bad, poor, fair, good, excellent), reflecting the annotation quality. For each dataset, we average the assessments of four experts and regard the output as the final rating. Although the video samples in Music\_video \cite{7_pandeya2021deep}—sourced from the music recordings—are more likely to elicit the emotional responses, and our eMotions presents unique annotation challenges, it still achieves the highest MOS score, as illustrated in Fig.~\ref{fig_MOS_Scores}, which highlights the superior consistency and reliability of our annotations compared to the existing datasets. These outcomes impressively substantiate the effectiveness of our efforts in augmenting labeling quality.

\subsection{Dataset Characteristics}
\label{subsec:3.5}

We summarize the statistical characteristics of our proposed eMotions dataset in Tab.~\ref{tab2}, detailing the number of videos, processed frames, and quantitative durations across six emotional categories. It is evident that the Excitation category contains the largest number of video samples, whereas the negative emotional classes (\textit{i.e.}, Fear, Sadness, Tension) contain relatively fewer video samples. Tab.~\ref{tab3} further outlines the distribution and ratio of raw and labeled video samples originating from three SVs platforms. We can figure out that the videos from Douyin and Kuaishou, two Chinese SVs platforms, account for the largest proportion.

\begin{table}[t!]
\centering
\setlength{\tabcolsep}{2pt}
\begin{center}
\setlength{\arrayrulewidth}{0.22pt}
\caption{The statistics of eMotions, detailing the number of videos and processed frames, along with the quantitative durations. The magnitude is $10^4$ for ``Total (s)" and ``Frames".}
\vspace{0.3em}
\label{tab2}
\small
\resizebox{\linewidth}{!}{%
\begin{tabular}{ccccccc}
\toprule[0.22pt]
Category & Videos & \begin{tabular}[c]{@{}c@{}}Total (s)\end{tabular} & \begin{tabular}[c]{@{}c@{}}Shortest (s)\end{tabular} & \begin{tabular}[c]{@{}c@{}}Longest (s)\end{tabular} & \begin{tabular}[c]{@{}c@{}}Average (s)\end{tabular} & \begin{tabular}[c]{@{}c@{}}Frames\end{tabular} \\ \midrule[0.22pt]
Excitation        & 11,739            & 29.35                                                              & 3.72                                                                      & 163.77                                                                   & 25.00                                                                    & 945.29                                                              \\
Fear              & 954              & 2.59                                                               & 2.81                                                                      & 117.49                                                                   & 27.08                                                                    & 78.56                                                               \\
Neutral           & 8,795             & 24.97                                                             & 2.46                                                                      & 150.93                                                                   & 28.39                                                                    & 815.11                                                              \\
Relaxation        & 2,214             & 5.24                                                              & 5.06                                                                      & 117.05                                                                   & 23.69                                                                    & 163.80                                                              \\
Sadness               & 2,090             & 4.04                                                              & 3.25                                                                      & 120.77                                                                   & 19.30                                                                    & 131.53                                                              \\
Tension           & 2,204             & 4.90                                                               & 3.79                                                                      & 119.32                                                                   & 22.25                                                                    & 152.15                                                              \\
\rowcolor{cvprblue!20}
Overall           & 27,996            & 71.09                                                              & 2.46                                                                      & 163.77                                                                   & 25.39                                                                    & 2,286.44                                                             \\ \hline
\end{tabular}%
}
\end{center}
\vspace{-0.35em}
\end{table}

\begin{table}[t!]
\centering
\renewcommand{\arraystretch}{0.95}
\setlength{\arrayrulewidth}{0.22pt}
\caption{The quantity and proportions of raw and labeled video samples across three SVs platforms.}
\vspace{0.2em}
\label{tab3}
\small
\resizebox{\linewidth}{!}{%
\begin{tabular}{cccccccc}
\toprule[0.22pt]
\multirow{2}{*}{\begin{tabular}[c]{@{}c@{}}Data \\Type\end{tabular}} & \multicolumn{2}{c}{Douyin} & \multicolumn{2}{c}{Kuaishou} & \multicolumn{2}{c}{Tiktok} & \multirow{2}{*}{Sum} \\ \cmidrule[0.22pt](lr){2-7}
                           & No.       & Ratio        & No.         & Ratio         & No.        & Ratio        &                        \\ \midrule[0.22pt]
Raw                   & 15,977       & 47.58\%       & 10,000        & 29.78\%        & 7,600        & 22.64\%       & 33,577                  \\
Labeled              & 12,395       & 44.27\%       & 8,264         & 29.52\%        & 7,337        & 26.21\%       & 27,996                  \\ \bottomrule[0.22pt]
\end{tabular}%
}
\vspace{-1.2em}
\end{table}

Fig.~\ref{fig_wordclouds_duration}~(a) and Fig.~\ref{fig_wordclouds_duration}~(b) detailedly illustrate the word clouds of topics and content types within the dataset. The results suggest that the predominant themes are closely related to everyday experiences, such as dancing and photography, as well as time-sensitive social issues including COVID-19 and natural disasters like floods. Content types are largely human-centric (\textit{e.g.}, daily life, livelihood, news). Moreover, there is a thematic alignment between certain topics and content categories in SVs (\textit{e.g.}, game, health). Figure~\ref{fig_wordclouds_duration}~(c) shows the distribution of video durations. A substantial portion of SVs (72.81\%) have durations between 0 and 30 seconds, underscoring their brevity and efficiency in catering to users’ demand for rapid information sharing on online web platforms. Furthermore, different from the human-centered VEA datasets (\textit{e.g.}, \cite{16_zadeh2018multimodal,26_busso2008iemocap,33_soujanya2018multimodal}), eMotions originating from SVs is directed towards in-the-wild scenes, implying that the emotional elicitation requires the whole duration, not short clips of video.

\subsection{Definitions of Two Variant Datasets}
\label{subsec:3.6}
To accommodate diverse research objectives, we construct two customized variant datasets from the original eMotions via targeted data sampling strategies. For eMotions\_balanced dataset, we randomly select samples from eMotions, in which 4,000 samples of Excitation, 3,000 samples of Neutral, and all the samples of the remaining categories. Regarding eMotions\_test, we randomly select the corresponding proportion of samples based on the ratio of each emotional category in eMotions, totaling 5,000 samples.

%% file: Sec/3d5_method.tex
\section{Methodology}
\label{sec:method}

\subsection{Visual and Audio Representations}
\label{subsec:4.1}
To tackle the inherent challenges of our proposed eMotions dataset, we learn audio-visual representations end-to-end. Unlike \cite{35_zhao2020end,40_zhang2023weakly}, we use Video Swin-Transformer (Video Swin-T) \cite{19_liu2022video} as the visual backbone for its ability to capture more semantically relevant features.
Specifically, let $\{(v_i, a_i)\}^b$ be a batch of $b$ samples, where $v_i$ and $a_i$ are the video and audio of the $i$-th sample.
For $v_i$, we split it into $s$ equal-duration segments, randomly selecting $T$ consecutive frames from each (preserving temporal order) as input snippets.
For each snippet $ v_i^{j} \in \mathbb{R}^{T \times H \times W \times 3}$, we first project it into patch features $ v_i^{j}\in \mathbb{R}^{\frac{T}{2} \times \frac{H}{4} \times \frac{W}{4} \times 96}$ (with $H, W$ as $v_i$'s height and width).
For the $l$-th stage in Video Swin-T, the input features can be represented as $F_v^{l} = \frac{T}{2} \times \frac{H}{2^{l+1}} \times \frac{W}{2^{l+1}} \times 2^{l-1}C $, $l \in \{1, 2, 3, 4\}$.
The first stage uses a linear embedding layer to project patch feature dimensions to $C$; remaining stages employ patch merging layers to progressively reduce feature map spatial size. Additionally, Video Swin-T blocks use the novel 3D SW-MSA Module to capture global relations and model spatial-temporal correlations. The final $v_i$ representations consist of snippet features $F_{v}(v_i) = \{f_{v}^{1}(v_i), f_{v}^{2}(v_i), ..., f_{v}^{s}(v_i)$\}, and $f_{v}^{j}(v_i) \in \mathbb{R}^{\frac{T}{2} \times \frac{H}{32} \times \frac{W}{32} \times 8C}$.

For the audio component $a_i$, we derive a continuous descriptor $a$ using MFCC. This descriptor $a$ is then center-cropped to a fixed length of $q$, with padding applied as needed to yield $a^\prime$. Subsequently, $a^\prime$ is split into $s$ snippets along its time-series axis through chunking and stacking. We employ ResNet34 \cite{22_he2016deep} to generate the final representations $F_{a}(a_i) = \{f_{a}^{1}(a_i), f_{a}^{2}(a_i), ..., f_{a}^{s}(a_i)\}$ and $ f_{a}^{p}(a_i) \in \mathbb{R}^{ H^\prime \times W^\prime \times C^\prime}$. Here, $H^\prime$, $W^\prime$, and $C^\prime$ correspond to the height, width, and final dimension, respectively.

Subsequently, we perform pooling across spatial-temporal dimensions, and then use fully connected layers to reshape $F_{v}$ and $F_{a}$ into $F_{v}$ and $F_{a}$ as $ F_{v} \in \mathbb{R}^{s \times C_{1}}$ and $ F_{a} \in \mathbb{R}^{s \times C_{1}}$, respectively.

\begin{figure*}[t!]
\setlength{\belowcaptionskip}{-1.5em}
\centering
\includegraphics[width=\textwidth]{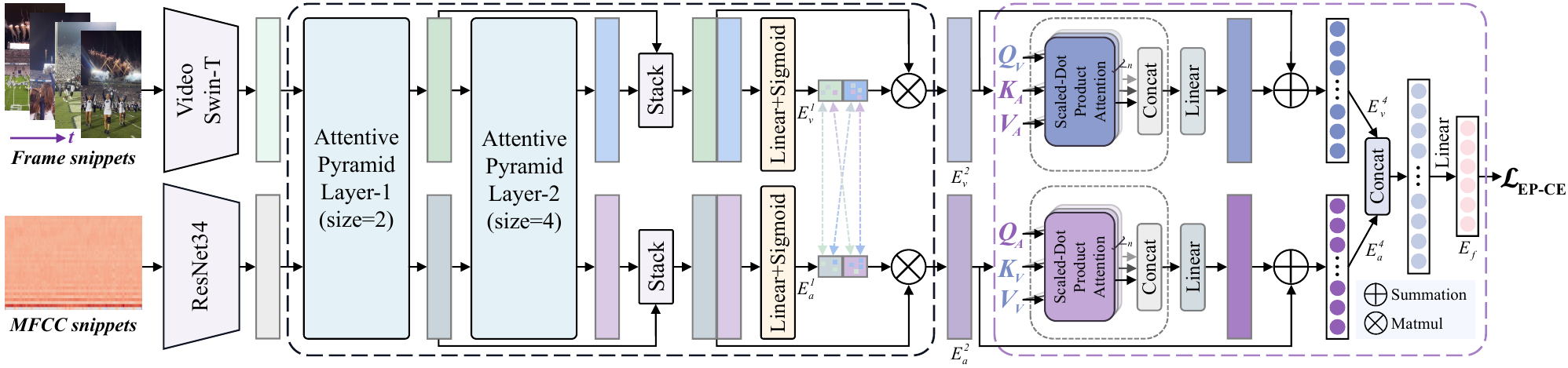}
\vspace{-0.6em}
\caption{The overall illustration of AV-CANet. Local-Global Fusion (LGF) Module consists of Local-level Interactive-Selective Fusion and Global-level Complementary Fusion Sub-Modules, framed by dashed lines in \textit{black} and \textit{\textcolor[RGB]{167,127,197}{purple}}, respectively.}
\label{fig5-AV-CANet}
\end{figure*}

\subsection{Local-Global Fusion Module}
\label{subsec:4.2-new}
We propose the LGF Module, which comprises the Local-level Interactive-Selective Fusion and Global-level Complementary Fusion Sub-Modules (referred to as LISF and GLCF). This module is designed to gradually capture the cross-modal correlations, thus alleviating the local biases and collective information gaps arising from emotional inconsistency.

Within the LISF Sub-Module, we first sequentially employ two attentive pyramid layers \cite{new_37_yu2022mm}, which are built on modified Self-Attention (SA) and Cross-Modal Attention (CMA) blocks, to enable locally dense interactions between audio and visual features. 
Specifically, to enable interactions across multiple contextual scales within the pyramid layers, a fixed interaction window $s_t(F,d) = [F_{t-d}, ..., F_{t+d}]$ is established. This is accomplished by applying masks to non-involved regions, where $d$ represents the size of the $t^{th}$ snippet $t \in [1, s]$, thereby confining the interaction scope of the blocks. Leveraging this fixed-size mechanism, the attention computations of SA and CMA blocks at the snippet level can be formulated as follows:
\begin{align}
\small
SA(F_{m},d) &= Att({F_{m}}W_{q}, s_t(F_{m},d)W_{k},  s_t(F_{m},d)W_{v}), \\
CMA(F_{m}, F_{\bar{m}}, d) &= Att({F_{m}}W_{q}, s_t(F_{\bar{m}},d)W_{k},  s_t(F_{\bar{m}},d)W_{v}), \\
Att(q, k, v) &= softmax(qk^T/\sqrt{d_k})v.
\label{eq-sa-cma-att}
\end{align}

Here, $q$, $k$, $v$, and $1/\sqrt{d_k}$ denote queries, keys, values, and the scaling factor, respectively. $W_{*}$ ($* \in \{q,k,v \}$ represents learnable parameters, and the modality $ m \in \{a, v\}$. Subsequently, the modified feed-forward layer, LayerNorm, and residual connection are utilized to construct the complete SA and CMA blocks.

In cases where the multiple attention heads are employed, the outputs generated by the SA and CMA blocks can be expressed as follows:
\begin{align}
F_{s}^m &= \mathbf{c}\left(Att_{sa}^1, ..., Att_{sa}^n\right)W_0, \\
F_{c}^m &= \mathbf{c}\left(Att_{cma}^1, ..., Att_{cma}^n\right)W_1.
\label{eq-sa-cma-mh}
\end{align}

Here, $\mathbf{c}(\cdot)$ and $n$ stand for concatenation and the number of attention heads, respectively, while $W_0$ and $W_1$ are learnable parameters. Additionally, the parameters within the CMA block are shared to enhance local-level interactions.

Upon obtaining the parallel outputs from the SA and CMA blocks, we first concatenate these outputs, and we compute channel-wise attention scores to refine the concatenated result. This refinement process involves passing the concatenated outputs through a linear layer followed by a sigmoid function. In the end, we perform summation to yield the local-aware audio-visual features:
\begin{equation}
\footnotesize
F_m^\prime = \sigma\left(W_{s} \cdot \mathbf{c}\left(F_{s}^m, F_{c}^m\right) + b_s\right)F_{s}^m + \sigma\left(W_{c} \cdot \mathbf{c}\left(F_{s}^m, F_{c}^m\right) + b_c\right)F_{c}^m,
\label{eq-overall}
\end{equation}
where $\mathbf{\sigma}(\cdot)$ denotes sigmoid function, $W_{*}$ and $b_{*}$ ($* \in \{s,c \}$) are the learnable parameters of linear layers.

Subsequent to the approaches presented in \cite{new_37_yu2022mm}, we utilize the dilated residual blocks to jointly extract temporal semantics and implement positional enhancements. Ultimately, the densely interactive outputs $\{F_v^i, F_a^i\}_{i=1}^{L}$ for the $t^{th}$ snippet are retained as pyramid-structured features, where $F_m^i \in \mathbb{R}^{1 \times C_{1}}$ and $L$ denotes the number of pyramid layers.

To dynamically perform selective integration of pyramid features at the snippet level, we introduce the Selectively Cross-Aggregative Integration Layer rather than merely using pooling operations. Specifically, for the $t^{th}$ snippet, we first stack $\{F_m^i\}_{i=1}^{L}$ along the dimension $L$ to combine interactive features from multiple pyramid layers. Then, linear projection and a sigmoid function are applied to dynamically assign weights to different granularities, which can be expressed as:
\begin{equation}
E_m^1 = \sigma\left(W_a \cdot s\left(F_m^1, ..., F_m^L\right)+b_a\right).
\end{equation}

In this context, $s(\cdot)$ denotes the stacking operation. Subsequently, we utilize the CMA block without a fixed-size window (where both the number of heads and the dimension are set to 1) to adaptively modulate the significance of audio-visual weights. Finally, a weighted summation of the pyramid layers is performed to yield the compatibly integrated features $E_m^2$. The mathematical formulation of the above operations is presented as follows:
\begin{equation}
E_m^2 = \sum_{l=1}^{L} s^{l}\left(F_m^1, ..., F_m^L\right) \cdot Att^{l}\left(E_m^1 W_q, E_{\bar{m}}^1 W_k, E_{\bar{m}}^1 W_v\right).
\end{equation}

At this time, our model has learned representations that are highly sensitive to local biases. To address the collective information gaps, we propose the GLCF Sub-Module, which enables the capture of correlated information from audio-visual features at the global level and generates the final outputs. Specifically, we first employ the globally unrestricted CMA block \cite{57_vaswani2017attention} with multiple attention heads. This block is used to complementarily learn audio-visual representations across different levels and capture varying degrees of correlations between snippets $s$. Following this, concatenation is carried out for non-local aggregations, that is:
\begin{equation}
\footnotesize
E_m^3 = \mathbf{c}\left(Att_{gca}^1, ..., Att_{gca}^i\left(E_m^2 W_q, E_{\bar{m}}^2 W_k, E_{\bar{m}}^2 W_v\right), ...,Att_{gca}^n\right) W_2.
\end{equation}

Subsequently, linear projection layers are employed to strengthen the emphasis on discriminative information and snippet-aware correlations. A residual connection is also incorporated to facilitate unimpeded information flow and gradient propagation. Following this, pooling layers are applied along the snippet dimension to reshape the features into $E_m^4 \in \mathbb{R}^{C_{2}}$:
\begin{equation}
E_m^4 = \delta\left(\left(W_{g} \cdot E_m^3 + b_{g}\right)+E_m^2\right).
\end{equation}

Here, $\delta(\cdot)$ represents average pooling. In the end, we employ the Mid-Concat fusion strategy, after which a category-specific linear layer is used to generate the final representations of our entire model, which can be formulated as:
\begin{equation}
E_f = W_{f} \cdot \mathbf{c}\left(E_m^4, E_{\bar{m}}^4\right) + b_{f}.
\end{equation}

\subsection{Emotion Polarity Enhanced CE Loss} 
\label{subsec:4.3}

Optimizing the conventional Cross-Entropy (CE) Loss directly may result in misclassification because of larger semantic disparities and challenges in learning emotion-related features. Taking inspiration from \cite{35_zhao2020end}, we introduce the Emotion Polarity-Cross-Entropy (EP-CE) Loss. This loss takes into account the \textit{neutral} polarity found in the eMotions dataset and numerous Valence-Arousal-Dominance (VED) datasets \cite{7_pandeya2021deep,18_lee2019context,26_busso2008iemocap}. Specifically, when the predicted emotion polarity varies from the actual emotion polarity, $\gamma_{\textit{ep}(y_i)}$ is used to assign weights to the model optimization process. The EP-CE Loss is defined as follows:
\vspace{-0.25em}
\begin{equation}
\mathcal{L}_{ep} = -\frac{1}{N} \sum_{i=1}^N \left(1 + \gamma_{\textit{ep}(y_i)} \cdot s(y_i, \hat{y_i})\right) \sum_{c=0}^{C-1} \beta_{[c=y_i]} \log p_{i, c}.
\label{eq11}
\vspace{-0.2em}
\end{equation}

In this context, $C$ denotes the total number of categories, $\beta_{\left[c=y_i\right]}$ serves as a binary indicator, and $p_{i, c}$ represents the predicted probability that sample $i$ falls into category \(c\). The term $c$. $\gamma_{\textit{ep}(y_i)}$ corresponds to the polarity coefficients that regulate the extent of penalties. Here, $y_i$ and $\hat{y_i}$ stand for the ground-truth and the prediction respectively. The function $s\left(y_i, \hat{y_i}\right)$ determines whether penalties should be added. Specifically, when $\operatorname{\textit{ep}}(y_i) \neq \operatorname{\textit{ep}}(\hat{y_i})$, $s\left(y_i, \hat{y_i}\right)$ is set to 1; otherwise, it is 0. The \(\textit{ep}(\cdot)\) function maps an emotional category to its corresponding polarity.
Moreover, to enhance the training stability when dealing with large-scale datasets, we initially implement the deferred update strategy after every 4 batches. Subsequently, we adopt a multi-task learning approach. In this approach, the losses from the overall model, as well as those from the visual and audio branches, are accumulated.

%% file: Sec/4_expers.tex
\section{Experiments}
\label{sec:experiment}

\begin{table*}[t!]
\centering
\begin{center}
\setlength{\arrayrulewidth}{0.22pt}
\caption{Performace comparisons of advanced VEA methods and AV-CANet on three eMotions-related datasets. We report the overall metrics of ACC, WA-F1, and UAR, as well as ACC of each emotional category. e\_B: eMotions\_balanced. e\_T: eMotions\_test. e\_All: eMotions. 3D-R50: 3D-ResNet50. R34: ResNet34. C53: CSP-DarkNet53. The separated columns from top to bottom are A, V, and A + V modalities, respectively. This paper highlights the best performance in \textbf{\underline{\textit{bold}}} and \underline{underline} the second performance.}
\vspace{0.05em}

\label{performace-comparisons}
\rule{\textwidth}{0pt}
\makebox[\textwidth][c]{
\resizebox{\textwidth}{!}
{
\renewcommand{\arraystretch}{1.5} 
\tabcolsep=0.6136pt
\fontsize{6.55}{7.2}\selectfont
\begin{tabular}{lllclcccccccccccccccccclccccccccc}
\hline
\multicolumn{2}{c}{\multirow{3}{*}{Method}}               &  & \multicolumn{18}{c}{ACC of Each Emotional Category (\%)}                                                                                                  &  & \multicolumn{9}{c}{Evaluation Metrics (\%)}                     \\
\cline{4-21} \cline{23-31} 
\multicolumn{2}{c}{}                                      &  & \multicolumn{3}{c}{EX} & \multicolumn{3}{c}{FR} & \multicolumn{3}{c}{NU} & \multicolumn{3}{c}{RE} & \multicolumn{3}{c}{SD} & \multicolumn{3}{c}{TN} &  & \multicolumn{3}{c}{ACC} & \multicolumn{3}{c}{WA-F1} & \multicolumn{3}{c}{UAR} \\ \cline{4-21} \cline{23-31}

\multicolumn{2}{c}{}                                      &  & e\_B  & e\_T  & e\_All & e\_B  & e\_T  & e\_All & e\_B  & e\_T  & e\_All & e\_B  & e\_T  & e\_All & e\_B  & e\_T  & e\_All & e\_B  & e\_T  & e\_All &  & e\_B   & e\_T  & e\_All & e\_B   & e\_T   & e\_All  & e\_B   & e\_T   & e\_All \\ \hline
   & ResNet34 \cite{22_he2016deep}                           &  & 66.63 & 70.00 & 78.92  & 50.79 & 41.18 & 37.17  & 37.67 & 57.32 & 55.66  & 68.62 & 51.90 & 44.70  & 37.08 & 12.00 & 16.27  & 52.38 & 17.95 & 22.90  &  & 53.44  & 55.20 & 58.39  & 52.65  & 53.13  & 56.14 & 52.19             & 41.72             & 42.60   \\
                     & PANNS \cite{27_kong2020panns}                              &  & 71.25 & 78.10 & \underline{83.48}  & 49.74 & 41.18 & 43.46  & 30.17 & 45.86 & 42.58  & 57.79 & 41.77 & 50.79  & 36.60 & 4.00  & 16.75  & 51.02 & 5.13  & 29.93  &  & 51.16  & 52.60 & 57.48  & 50.06  & 48.48  & 54.99  & 49.43             & 36.01             & 44.50 \\
                     & ECAPA-TDNN \cite{new_32_desplanques2020ecapa}                         &  & 69.13 & \underline{80.24} & \underline{\textbf{84.63}}  & 42.93 & 29.41 & 29.32  & 36.83 & 39.81 & 47.24  & 49.21 & 31.65 & 42.66  & 55.02 & 36.00 & 23.68  & 38.78 & 14.10 & 13.38  &  & 50.99  & 53.50 & 57.52  & 50.45  & 50.92  & 54.55 & 48.65             & 38.53            & 40.15  \\
                     & Res2Net \cite{34_gao2019res2net}                            &  & \underline{\textbf{73.63}} & 75.24 & 71.98  & 43.98 & 32.35 & 40.84  & 35.83 & 47.13 & 51.51  & 57.79 & 45.57 & 48.98  & 26.08 & 5.33  & 17.70  & 44.44 & 20.51 & 30.39  &  & 50.09  & 53.10 & 55.34  & 48.72  & 50.37  & 53.96  & 46.96             & 37.69             & 43.57 \\
                     & CAMPPlus \cite{38_wang2023cam++}                           &  & 69.00 & 73.03 & 78.62  & 40.31 & 41.18 & 34.56  & 36.50 & 48.09 & 49.91  & 65.46 & 51.90 & 48.08  & 38.28 & 10.67 & 13.88  & 48.53 & 25.32 & 18.14  &  & 52.26  & 54.00 & 56.09  & 51.37  & 52.06  & 53.51 & 49.68              & 41.70             & 40.53  \\ \hline

 & Video Swin-Transformer \cite{19_liu2022video}             &  & 66.67 & 66.83 & 65.64  & 66.14 & \underline{61.76} & 49.21  & \underline{\textbf{69.76}} & \underline{80.25} & \underline{\textbf{86.80}}  & 71.37 & 56.96 & 55.30  & 51.63 & 40.00 & 40.19  & 55.03 & 11.39 & 10.20  &  & 64.12  & 63.70 & 64.64  & \underline{64.31}  & 62.21  & 62.92 & 63.43             & 52.87            & 51.23  \\
                     & I3D \cite{43_carreira2017quo}                                &  & 69.84 & 68.74 & 71.48  & 74.87 & \underline{\textbf{64.71}} & 52.36  & 62.83 & \underline{\textbf{80.89}} & 80.66  & 70.20 & \underline{\textbf{59.49}} & 51.92  & 51.67 & 37.33 & 42.58  & 51.02 & 6.33  & 13.15  &  & 63.28  & \underline{64.40} & 65.41  & 63.68  & \underline{62.68}  & 63.90 & 63.41             & \underline{52.91}             & 52.03  \\
                     & SlowFast \cite{new_33_feichtenhofer2019slowfast}                           &  & 70.34 & 76.37 & 77.71  & 5.76  & 0.00  & 0.00   & 22.33 & 47.77 & 51.42  & 38.37 & 7.59  & 15.35  & 45.93 & 14.67 & 11.24  & 48.98 & 0.00  & 3.40   &  & 44.43  & 48.70 & 51.05  & 41.94  & 42.77  & 45.60  & 38.62             & 24.40             & 26.52 \\
                     & CTEN \cite{40_zhang2023weakly}                              &  & 65.00   & 74.22 &73.51        &37.17       & 38.24 &28.80        &45.83       & 61.46 &65.87        &69.07       & 54.43 &45.82        &52.15       & 32.00 &30.62        &36.28       & 7.59  &6.12        &  &53.58        & 59.00 &58.89        &53.02        & 56.94  &56.53    & 50.92                 & 44.66             & 41.79     \\
                     & Former-DFER \cite{new_34_zhao2021former}                        &  & 64.85 & 74.46 & 69.97  & 41.80 & 38.24 & 39.79  & 36.01 & 50.32 & 65.55  & 58.57 & 26.58 & 51.69  & 43.36 & 24.00 & 39.71  & 51.68 & 26.58 & 10.43  &  & 51.64  & 54.30 & 59.16  & 51.38  & 52.56  & 57.67 & 49.38             & 40.03             & 46.19  \\
                     & TimeSformer \cite{30_bertasius2021space}                        &  & 68.84 & 70.88 & 78.90  & 74.35 & 55.88 & 51.31  & \underline{63.33} & 75.48 & 73.09  & 73.81 & 40.51 & 47.18  & 57.89 & 28.00 & 34.45  & 48.75 & 8.86  & 4.31   &  & \underline{64.18}  & 61.30 & 64.43  & \underline{\textbf{64.32}}  & 59.17  & 61.62  & 64.50             & 46.60             & 48.21 \\
                     & DFER-CLIP \cite{new_36_zhao2023prompting}                             &  & 59.40 & 67.48 & 69.55   & \underline{\textbf{84.29}} & 50.00 & \underline{\textbf{92.38}}   & 45.49  & 53.44  & 57.12   & \underline{74.32}  & 57.14  & \underline{\textbf{73.56}}   & 63.98  & \underline{44.68}  & \underline{\textbf{67.60}}   & 51.42  & \underline{39.13}  & \underline{40.80}   &  & 58.82   & 58.90  & 64.33   & 59.04   & 56.96  & 63.07  & 58.41                 & 39.75                 & 51.49  \\
                     & C3D \cite{32_tran2015learning}                                &  & \underline{72.84} & 76.61 & 65.94  & 69.63 & 55.88 & 41.88  & 51.33 & 62.74 & \underline{82.76}  & 69.53 & 49.37 & 58.92  & 55.50 & 37.33 & 43.78  & 51.25 & 12.66 & 3.40   &  & 61.86  & 61.40 & 63.27  & 61.84  & 59.72  & 61.04  & 61.68             & 49.10             & 49.45 \\ \hline

& VAANet \cite{35_zhao2020end}                             &  & 72.38 & 
                     74.46 & 72.79  & 63.35 & 47.06 & 58.64  & 47.83 & 57.01 & 65.43  & 67.95 & 54.43 & 59.14  & 44.50 & 28.00 & 45.22  & 59.41 & 37.97 & 32.88  &  & 60.01  & 60.10 & 63.71  & 59.89  & 59.46  & 63.31  & 59.24             & 49.82             & 55.68 \\
                     & OGM-GE \cite{50_peng2022balanced}                             &  & 67.27 & 77.00 & 80.32  & 67.20 & 35.29 & 34.34  & 33.92 & 48.30 & 42.71  & 46.64 & 27.69 & 39.62  & 31.33 & 26.83 & 18.33  & 56.60 & 8.57  & 34.20  &  & 50.78  & 54.20 & 55.39  & 50.02  & 51.49  & 53.19  & 50.49             & 37.28             & 41.59 \\
                     & MSAF \cite{new_35_su2020msaf}                             &  & 60.33  & 63.43  & 71.57   & 54.77  & 44.83  & 51.32   & 41.30  & 60.52  & 65.06   & 70.95  & 37.18  & 60.92   & 52.36  & 26.39  & 37.03   & 47.80  & 15.28   & 12.33   &  & 54.69   & 53.73  & 60.93   & 54.51   & 52.70   & 59.51  & 54.58                 & 41.27                 & 49.70  \\
                     & MMCosine \cite{53_xu2023mmcosine}                          &  & 59.03 & 78.17 & 68.93  & 35.45 & 23.53 & 48.99  & 41.08 & 45.51 & 53.87  & 59.22 & 30.77 & 43.63  & 57.89 & 28.05 & 34.29  & 50.34 & 7.14  & 23.53  &  & 52.47  & 53.60 & 55.27  & 52.46  & 50.60  & 54.35 & 50.50             & 35.53             & 45.54  \\
                     & 3D-R50 + R34 \cite{21_hara2018can,22_he2016deep}             &  & 72.50 & 68.02 & 77.73  & 69.63 & 35.29 & 60.21  & 49.17 & 69.43 & 59.24  & 72.69 & \underline{58.23} & 60.50  & 55.74 & 36.00 & 36.36  & 55.78 & 17.72 & \underline{\textbf{41.04}}  &  & 62.53  & 60.20 & 63.98  & 62.31  & 59.04  & 63.45 & 62.59             & 47.45             & 55.85  \\ 
                     
                     & ConvNeXt + C53 \cite{46_liu2022convnet,42_wang2020cspnet} &  & 66.25 & 77.33 & 78.36  & \underline{75.92} & 50.00 & \underline{66.49}  & 46.17 & 62.74 & 61.46  & 69.30 & 50.63 & 57.11  & \underline{\textbf{65.55}} & 34.67 & \underline{54.55}  & \underline{\textbf{65.08}} & 35.44 & 33.11  &  & 62.91  & 63.20 & \underline{65.62}  & 62.87  & 62.33  & \underline{65.01} & \underline{64.71}             & 51.80            & \underline{58.51}  \\ 
                     
                     \rowcolor{cvprblue!20}
                     & AV-CANet                           &  & 67.03   & \underline{\textbf{82.58}}   & 80.24    & 68.25   & 58.82   & 59.16    & 50.87   &62.10       & 63.39        & \underline{\textbf{75.05}}       & 45.57  & \underline{63.43}        & \underline{65.16}   & \underline{\textbf{48.00}}       & 54.07       & \underline{63.53}       & \underline{\textbf{46.84}}   & 40.14    &  & \underline{\textbf{64.40}}        & \underline{\textbf{67.00}}       & \underline{\textbf{67.79}}        & \underline{\textbf{64.32}}        & \underline{\textbf{66.28}}        & \underline{\textbf{67.32}} & \underline{\textbf{64.99}}                       & \underline{\textbf{57.32}}                       & \underline{\textbf{60.07}}         \\ \hline
\end{tabular}
}
}
\end{center}
\vspace{-0.5em}
\end{table*}

\begin{table*}[!ht]
\centering
\setlength{\arrayrulewidth}{0.22pt}
\caption{The overall performace comparisons of recent baseline methods and AV-CANet on Ekman6 (Ek6), VideoEmotion8 (Ve8), Music\_video (Mv), and IEMOCAP (IE). EM: Evaluation metrics. *: Lab-controlled dataset. -: Unavailable results.}
\vspace{0.05em}
\label{tab-performance-comparisons-public-datasets}
\noindent
\rule{\linewidth}{0pt}
\resizebox{\textwidth}{!}
{
\small
\renewcommand{\arraystretch}{1.5} 
\tabcolsep=1.67pt
\begin{tabular}{cclccccclccccccclcccccccclc}
\hline
\multirow{2}{*}{Dataset} & \multirow{2}{*}{EM} &  & \multicolumn{5}{c}{A}                                &  & \multicolumn{8}{c}{V}                                                    &  & \multicolumn{8}{c}{A+V}                                                                      \\ \cline{4-8} \cline{10-17} \cline{19-27}
                         &                         &  & \cite{22_he2016deep} & \cite{27_kong2020panns} & \cite{new_32_desplanques2020ecapa} & \cite{34_gao2019res2net}  & \cite{38_wang2023cam++} &  & \cite{19_liu2022video} & \cite{43_carreira2017quo} & \cite{new_33_feichtenhofer2019slowfast} & \cite{40_zhang2023weakly}  & \cite{new_34_zhao2021former} & \cite{30_bertasius2021space} & \cite{new_36_zhao2023prompting} & \hspace{0.05cm}\cite{32_tran2015learning} &   & \cite{35_zhao2020end} & \cite{50_peng2022balanced} & \cite{7_pandeya2021deep} & \cite{new_35_su2020msaf} & \cite{53_xu2023mmcosine} & \cite{21_hara2018can,22_he2016deep} & \cite{46_liu2022convnet,42_wang2020cspnet} &  & \textbf{Ours}             \\ \hline
\multirow{2}{*}{Ek6}     & ACC                     &  & 33.95    & 27.47    & 28.40    & 31.17    & 29.01    &  & 47.59    & 40.51   & 28.35    & \underline{53.44}    & 37.96    & 36.33   & 44.44 & 44.05    &   & 50.31    & 34.26    & -      & 30.56      & 33.33    & 48.44       & 50.00       &  & \underline{\textbf{55.63}}              \\
                         & WA-F1                   &  & 32.17    & 21.77    & 28.21    & 29.64    & 29.03    &  & 49.44    & 42.10   & 21.92    & \underline{52.76}    & 36.88    & 34.34   & 43.81 & 42.84    &  & 49.65    & 33.10    & -      & 25.35      & 33.08    & 47.57       & 49.33       &  & \underline{\textbf{55.44}}              \\ \hline
\multirow{2}{*}{Ve8}     & ACC                     &  & 31.63    & 27.91    & 26.98    & 29.77    & 22.33    &  & 46.31    & 40.89   & 25.12    & \underline{47.66}    & 35.35    & 46.30   & 37.21 & 42.36    &   & 46.73    & 33.33    & -      & 31.25      & 32.18    & 44.39       & 45.33       &  & \underline{\textbf{49.53}}                       \\
                         & WA-F1                   &  & 29.11    & 23.06    & 23.20    & 27.12    & 23.59    &  & 45.77    & 41.34   & 16.06    & \underline{47.04}    & 29.10    & 45.65   & 29.32 & 42.02    &  & 43.92    & 32.12    & -      & 28.56      & 26.82    & 43.21       & 45.23       &  & \underline{\textbf{48.24}}                   \\ \hline
\multirow{2}{*}{Mv}      & ACC                     &  & 72.73    & 63.38    & 69.95    & 65.91    & 70.71    &  & 74.68    & 64.05   & 50.13    & 76.20    & 60.51    & 78.23   & 79.49 & 62.78    &  & 78.99    & 57.32    & \underline{83.30}    & 63.38      & 58.59    & 73.42       & 75.70       &  & \underline{\textbf{84.81}}                    \\
                         & WA-F1                   &  & 73.07    & 61.46    & 69.95    & 65.76    & 71.79    &  & 80.64    & 79.75   & 45.17    & 75.43    & 59.73    & 81.07   & 79.51 & 77.43    &  & 77.99    & 56.46    & \underline{84.00}    & 63.06      & 57.70    & 72.98       & 74.25       &  & \underline{\textbf{84.63}}                    \\ \hline
\multirow{2}{*}{IE*}    & ACC                     &  & 53.09    & 53.44    & 55.31    & 48.78    & 51.81    &  & 68.48    & 68.64   & 27.54    & 55.31    & 54.26    & 38.27   & 32.91 & \underline{\textbf{68.96}}    &  & 62.08    & 49.07    & -      & 50.70      & 49.42    & 63.36       & 61.03       &  & \underline{68.73}                    \\
                         & WA-F1                   &  & 52.45    & 52.83    & 55.17    & 47.36    & 50.88    &  & 67.44    & 67.99   & 16.67    & 54.96    & 54.18    & 31.81   & 16.29 & \underline{68.16}    &  & 61.86    & 48.82    & -      & 49.76      & 49.42    & 63.12       & 60.26       &  & \underline{\textbf{68.73}}                    \\ \hline
\end{tabular}
}
\vspace{-0.8em}
\end{table*}

\subsection{Experimental Settings}

The entire framework is primarily constructed on PyTorch 1.8.1+cu101, leveraging NVIDIA RTX 3090 GPUs. Following \cite{19_liu2022video}, the AdamW optimizer \cite{65_loshchilov2017decoupled} is employed, with a weight decay of 2$\times10^{-2}$ and a learning rate ($lr$) of 1.5$\times10^{-4}$. All the proposed models undergo training for 30 epochs, using a batch size of 8, to guarantee fair comparisons. Both the visual and audio branches are initialized with 2D weights pre-trained on ImageNet \cite{59_deng2009imagenet}.
For the visual branch, the input is divided into 16 segments, from which 8 frames are randomly selected as input snippets. Each frame is then resized such that the length of its short side is 112 pixels. Data augmentation techniques mainly involve random horizontal flips and random cropping of size 112 $\times$ 112.
Concerning the audio branch, the audio is first converted to monophonic format, and the sampling rate is set at 44,100 Hz. Then, the audio is transformed into Numpy arrays and normalized. After that, the \href{https://librosa.org}{Librosa} library is utilized to calculate MFCC, with 32 coefficients. Subsequently, the processed audio is split into 16 snippets to serve as the input for the audio branch. The dimensions $C_1$ and $C_2$ are both set to 512.
The samples in the eMotions dataset are randomly shuffled and divided into non-overlapping training (80\%) and testing (20\%) sets. In accordance with standard practices \cite{40_zhang2023weakly,73_chudasama2022m2fnet,35_zhao2020end}, Accuracy (ACC), Weighted Average F1-Score (WA-F1), and Unweighted Average Recall (UAR) are employed as the global evaluation metrics.

\subsection{Main Results}
\noindent \textbf{Performance Comparisons on Three eMotions-related Datasets.} 
To validate the efficacy of our proposed AV-CANet, we extensively conduct comparisons with 19 baseline methods across three eMotions-related datasets. These baselines cover three modality types: audio-only (A), visual-only (V), and audio-visual fusion (A+V), with detailed results presented in Tab.~\ref{performace-comparisons}. The key findings are summarized as follows: \textbf{(1)} Visual features play a dominant role in audio-visual video emotion analysis (VEA), and in general, more advanced visual architectures tend to yield greater performance gains. \textbf{(2)} Compared with other methods, AV-CANet demonstrates relatively balanced performance across the six emotion categories on all three datasets. This suggests that it can proficiently interpret emotions in short videos (SVs). \textbf{(3)} Owing to its effective design that addresses the specific challenges of eMotions, AV-CANet achieves superior performance across all the metrics. To be specific, compared to baseline VAANet~\cite{35_zhao2020end}, it delivers improvements of 4.08\% ACC, 4.01 WA-F1, and 4.39 UAR on the overall eMotions dataset.

\noindent \textbf{Performance Comparisons on Public Datasets.}~In Tab.~\ref{tab-performance-comparisons-public-datasets}, the comparative results on four public VEA datasets show that our model performs favorably against recent VEA methods of different modalities, demonstrating its generalizability and robustness for various application-oriented VEA. Additionally, considering the observations in model performance across datasets, some VEA-oriented methods underperform the fine-tuned general models with large-sacle pre-training, implying the importance of large-scale datasets for VEA \cite{new_45_sun2024hicmae,new_52_sun2023mae}.

\begin{table}[]
\centering
\renewcommand{\arraystretch}{1.05} 
\setlength{\arrayrulewidth}{0.22pt}
\caption{Performance comparisons of our LGF Module with advanced overall fusion methods.}
\vspace{0.05em}
\small
\resizebox{\linewidth}{!}{%
\begin{tabular}{ccccccc}
\toprule[0.22pt]
\multirow{2}{*}{\raisebox{-2ex}{\begin{tabular}[c]{@{}c@{}}Methods\end{tabular}}} & \multicolumn{2}{c}{eMotions}    & \multicolumn{2}{c}{Music\_video} & \multicolumn{2}{c}{Ekman6}      \\ \cmidrule[0.22pt](l){2-3}   \cmidrule[0.22pt](l){4-5}   \cmidrule[0.22pt](l){6-7} 
                                                                            & ACC         & WA-F1          & ACC         & WA-F1          & ACC         & WA-F1          \\ \midrule[0.22pt]
Vanilla                                                                & 65.98          & 65.05          & 75.19          & 74.66 & 50.31          & 50.41          \\
Self-Attention \cite{57_vaswani2017attention}                                                                & \underline{67.21}          & \underline{66.72}          & 81.77          & 81.56 & 51.56          & 50.73          \\
LGI-Former \cite{new_52_sun2023mae}          & 67.13          & 66.51          & \underline{82.03}          & \underline{82.08}          & \underline{53.75}          & \underline{53.20}          \\
Sun et al. \cite{new_45_sun2024hicmae}                                                         & 66.14          & 65.15          & 77.97          & 77.62          & 49.69         & 49.29          \\
\rowcolor{cvprblue!20}
LGF Module                                                                   & \textbf{\underline{67.48}} & \textbf{\underline{67.19}} & \textbf{\underline{82.78}} & \textbf{\underline{82.46}}          & \textbf{\underline{54.06}} & \textbf{\underline{53.51}} \\ 
\hline
\end{tabular}
}
\label{tab-fusion-compare}
\end{table}

\begin{table}[t!]
\centering
\setlength{\arrayrulewidth}{0.22pt}
\caption{Performance comparisons of popular MLLMs and our AV-CANet in terms of ACC and WA-F1 on e\_T.}
\vspace{0.05em}
\small
\renewcommand{\arraystretch}{0.95} 
\resizebox{0.7\linewidth}{!}{%
\begin{tabular}{cccccc}
\hline
Method & ACC & WA-F1 \\ \hline
LLaVA~\cite{new_61_LLaVA_liu2024visual}          & 49.06  & 47.92   \\
Qwen-VL~\cite{new_62_qwenvl_bai2023qwen}         & 49.33  & 50.02   \\
InternVL~\cite{new_63_chen2024internvl}          & 52.54  & 52.88   \\
Video-LLaMA~\cite{new_64_video-llama-zhang2023video} & 56.02  & 55.11   \\
\rowcolor{cvprblue!20}
AV-CANet (ours)                                         & \textbf{67.00}  & \textbf{66.28} \\ \hline
\end{tabular}
}
\label{MLLM_compare}
\end{table}

\begin{figure}[t!]
\setlength{\belowcaptionskip}{-0.5cm}
\centering
\includegraphics[width=\linewidth]{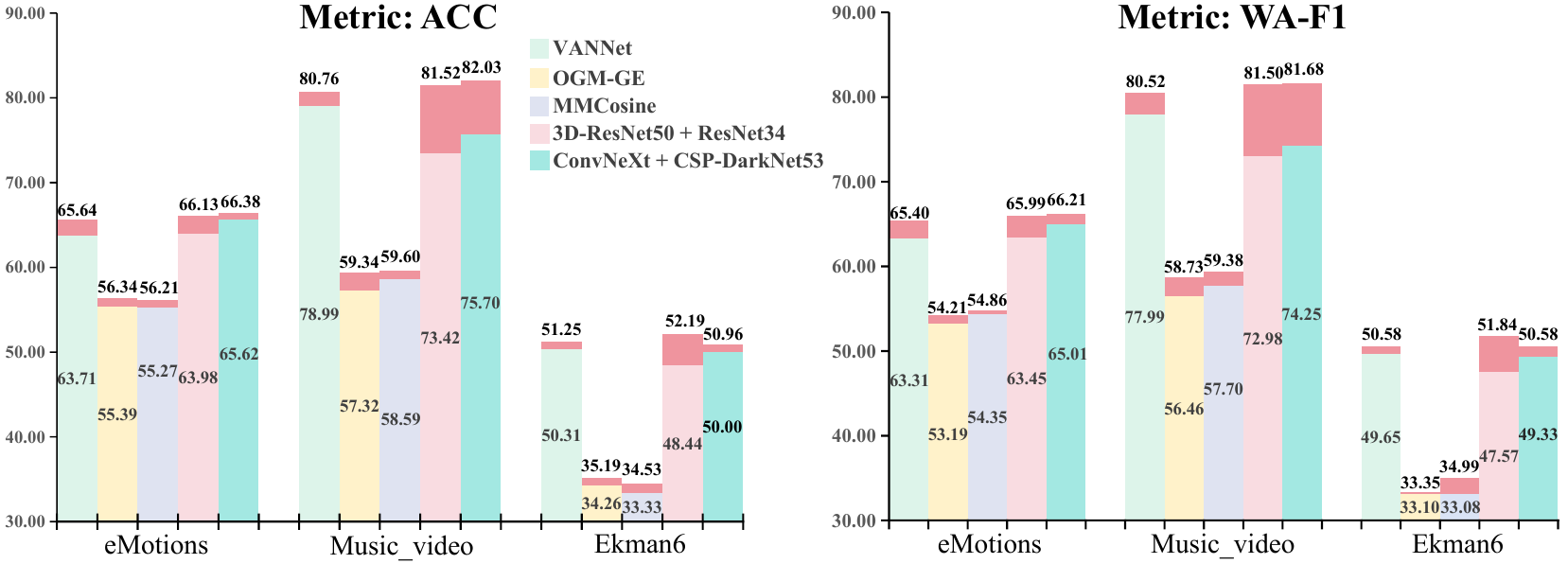}
\vspace{-1.0em}
\caption{The comparative results of five baselines adopting our LGF Module with their original performance. \textcolor[RGB]{239,148,158}{\textit{Pink}} is utilized to highlight performance gains.}
\label{fig_perform-compare-LGF}
\end{figure}

\noindent \textbf{Performance comparisons of MLLMs and our AV-CANet.} Multi-modal large language models (MLLMs)~\cite{new_61_LLaVA_liu2024visual,new_62_qwenvl_bai2023qwen,new_63_chen2024internvl} have demonstrated impressive performance on various downstream tasks. Inspired by this observation, we have adopted several mainstream MLLMs on our eMotions\_test (e\_T) dataset to explore their effectiveness and generalization capabilities on VEA, as shown in Tab~\ref{MLLM_compare}. From these extensive experimental results, we can figure out that compared to our targetedly designed AV-CANet, current MLLMs still have performance gaps, indicating that strong alignment and supervised fine-tuning (SFT) are very necessary for MLLMs to exhibit significant outcomes on VEA for SVs. In our future development, we plan to purposefully introduce strong MLLMs and design targeted strategies to improve overall performance.

\noindent \textbf{Effectiveness of LGF Module.}

As displayed in Tab.~\ref{tab-fusion-compare} above, we fairly compare our LGF Module with advanced overall fusion methods employing the same backbones on three datasets. We can clearly figure out that the LGF Module consistently outperforms these methods, with leading margins of up to 0.75\% ACC and 0.38 WA-F1, verifying its effectiveness.
Moreover, we replace the fusion modules of five audio-visual baselines with the LGF Module (Fig.~\ref{fig_perform-compare-LGF}). Results show that integrating the LGF Module consistently improves performance across three datasets, with maximum gains of 8.10\% ACC and 8.52 WA-F1, further confirming its superiority.

\subsection{Ablation Studies}

\begin{figure}[t!]
\setlength{\belowcaptionskip}{-0.5cm}
\centering
\includegraphics[width=\linewidth]{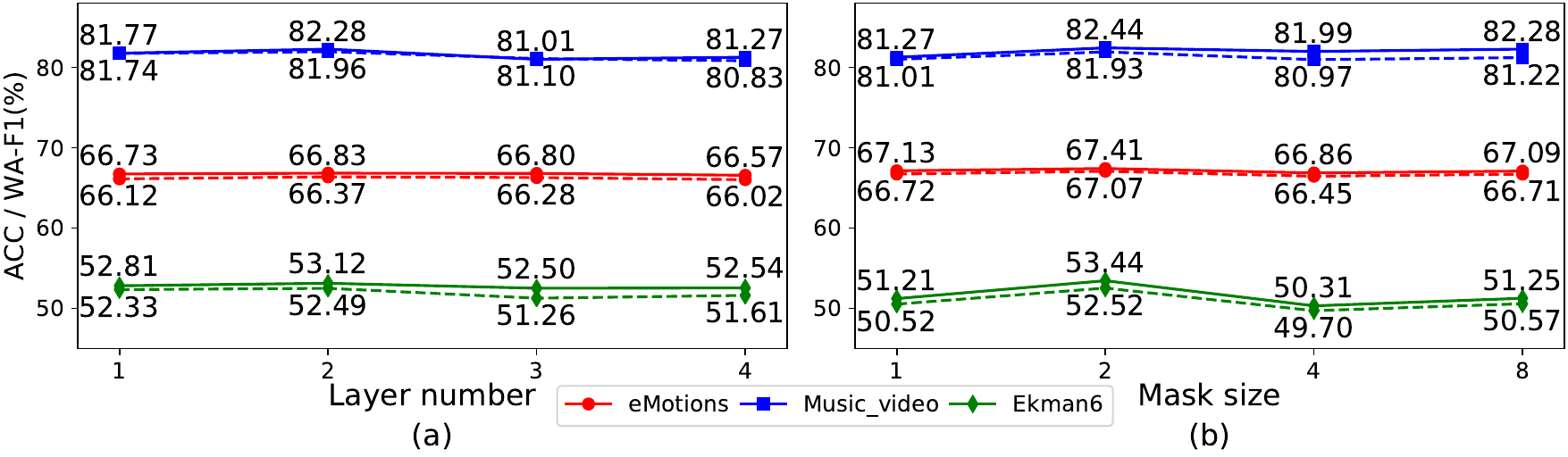}
\vspace{-1.5em}
\caption{Ablation study in investigating the influence of different layer number and mask sizes on LGF Module. The solid and dashed lines denote ACC and WA-F1, respectively.}
\label{fig-hryper}
\vspace{1.0em}
\end{figure}

We conduct systematic ablation studies on three datasets to analyze key factors of AV-CANet, with findings as follows: \textbf{(1)} As displayed in Fig.~\ref{fig-hryper} (a), testing different layer counts in the LISF Sub-Module reveals that pyramid layer number is not positively correlated with performance—excessive layers cause overly dense local interactions, increasing complexity and instability. \textbf{(2)} Fig.~\ref{fig-hryper} (b) demonstrates that improper mask sizes reduce model performance, highlighting the AV-CANet’s sensitivity to this hyperparameter. This result suggests that fine-grained tuning of mask-related settings is necessary to fully exploit the model’s potential. \textbf{(3)} Evaluating fusion strategies in the GLCF Sub-Module (Tab.~\ref{tab-fusion-strategies}) shows Mid-Concat yields the largest gains (0.94\% Acc, 1.02 WA-F1) across datasets, while EW-Multiply and Sum lead to declines, highlighting the critical role of strategy adaptability. \textbf{(4)} Ablating the two LGF sub-modules (Tab.~\ref{tab-LISF-GLCF}) demonstrates their joint use achieves the highest improvement, validating their superiority in modeling audio-visual correlations. \textbf{(5)} As shown in Tab.~\ref{tab-ep-ce-loss}, $\gamma_{\textit{ep}(y_i)}$ = 0.7 yields optimal performance across datasets. Larger penalties help the model focus on misclassified samples and enhance emotion-related representation learning, while smaller penalties on large-scale datasets may confuse learning, causing slight drops.

\begin{table}[t!]
\centering
\renewcommand{\arraystretch}{0.90} 
\setlength{\arrayrulewidth}{0.22pt}
\caption{The ablation comparisons of five different fusion strategies in the GLCF Sub-Module. EW: Element-Wise.}
\vspace{0.05em}
\label{tab-fusion-strategies}
\small
\resizebox{\linewidth}{!}{%
\begin{tabular}{ccccccc}
\toprule[0.22pt]
\multirow{2}{*}{\raisebox{-2ex}{\begin{tabular}[c]{@{}c@{}}Fusion \\Strategy\end{tabular}}} & \multicolumn{2}{c}{eMotions}    & \multicolumn{2}{c}{Music\_video} & \multicolumn{2}{c}{Ekman6}      \\ \cmidrule[0.22pt](l){2-3}   \cmidrule[0.22pt](l){4-5}   \cmidrule[0.22pt](l){6-7} 
                                                                            & ACC         & WA-F1          & ACC         & WA-F1          & ACC         & WA-F1          \\ \midrule[0.22pt]
Gated \cite{60_kiela2018efficient}                                                                & 67.39          & \textbf{67.35}          & 80.51          & 80.26 & 50.94          & 50.45          \\
EW-Multiply          & 64.44          & 63.97          & 76.71          & 76.44          & 48.44          & 47.17          \\
Neural                                                                & 66.52          & 66.12          & 82.03          & 81.75          & 53.44          & 52.34          \\
\rowcolor{cvprblue!20}
Mid-Concat                                                         & \textbf{67.48}          & 67.19          & \textbf{82.78}          & \textbf{82.46}          & \textbf{54.06}          & \textbf{53.51}          \\
Sum                                                                   & 66.49 & 66.17 & 81.77 & 81.64          & 50.94 & 50.18 \\ \bottomrule[0.22pt]
\end{tabular}
}
\vspace{-0.3em}
\end{table}

\begin{table}[t!]
\centering
\setlength{\abovecaptionskip}{-0em} 
\renewcommand{\arraystretch}{0.90}
\setlength{\arrayrulewidth}{0.22pt}
\caption{Ablations for the LISF and GLCF Sub-Modules.}
\small
\resizebox{\linewidth}{!}{%
\begin{tabular}{ccccccc}
\toprule[0.22pt]
\multirow{2}{*}{\makecell{LG-CF \\ Module}} & \multicolumn{2}{c}{eMotions} & \multicolumn{2}{c}{Music\_video} & \multicolumn{2}{c}{Ekman6} \\ 
\cmidrule[0.22pt](l){2-3}   \cmidrule[0.22pt](l){4-5}   \cmidrule[0.22pt](l){6-7} 
                                  & ACC        & WA-F1        & ACC          & WA-F1         & ACC       & WA-F1       \\ \midrule[0.22pt]
None                         & 65.98         & 65.05        & 75.19           & 74.66         & 50.31        & 50.41       \\
LISF                        & 67.41         & 67.07        & 82.44           & 81.93         & 53.44        & 52.52       \\
GLCF                      & 67.08         & 66.41        & 81.89           & 81.00         & 51.88        & 51.30       \\
\rowcolor{cvprblue!20}
LISF + GLCF                       & \textbf{67.48}         & \textbf{67.19}        & \textbf{82.78}           & \textbf{82.46}         & \textbf{54.06}        & \textbf{53.51}       \\ \hline
\end{tabular}
}
\label{tab-LISF-GLCF}
\vspace{-0.2em}
\end{table}

\begin{table}[t!]
\centering
\renewcommand{\arraystretch}{0.90} 
\setlength{\arrayrulewidth}{0.22pt}
\caption{The ablation comparisons of five different penalties for EP-CE Loss. Note that $\gamma_{neu}$ doesn't function on the Ekman6 dataset.}
\vspace{0.05em}
\label{tab-ep-ce-loss}
\small
\resizebox{\linewidth}{!}{%
\begin{tabular}{ccccccc}
\toprule[0.22pt]
\multirow{2}{*}{$\gamma_{pos}$: $\gamma_{neu}$: $\gamma_{neg}$} & \multicolumn{2}{c}{eMotions} & \multicolumn{2}{c}{Music\_video} & \multicolumn{2}{c}{Ekman6} \\ 
\cmidrule[0.22pt](l){2-3}   \cmidrule[0.22pt](l){4-5}   \cmidrule[0.22pt](l){6-7} 
                                  & ACC        & WA-F1        & ACC          & WA-F1         & ACC       & WA-F1       \\ \midrule[0.22pt]
0.3 : \textcolor{gray}{0.3} : 0.3                         & 67.34         & 66.83        & 84.03           & 83.75         & 54.37        & 54.18       \\
0.4 : \textcolor{gray}{0.4} : 0.4                        & 67.34         & 66.89        & 84.05           & 83.89         & 54.38        & 53.88       \\
0.5 : \textcolor{gray}{0.5} : 0.5                       & 67.53         & 67.22        & 84.81           & 84.58         & 55.00        & 54.29       \\
\rowcolor{cvprblue!20}
0.7 : \textcolor{gray}{0.7} : 0.7                       & \textbf{67.79}         & \textbf{67.32}        & \textbf{84.81}           & \textbf{84.63}         & \textbf{55.63}        & \textbf{55.44}       \\ \hline
\end{tabular}
}
\vspace{-1.0em}
\end{table}

\begin{table*}[t!]
\begin{center}
\caption{The performace comparisons of our AV-CANet using the initialization weights from ImageNet \cite{59_deng2009imagenet} and eMotions on Ekman6, VideoEmotion8, Music\_video, and IEMOCAP. $\dagger$: AV-CANet is initialized using the weight learned from eMotions.} 
\vspace{0.05em}
\label{cross-dataset}
\renewcommand{\arraystretch}{1.15} 
\setlength{\tabcolsep}{10pt}
\resizebox{\linewidth}{!}{
\begin{tabular}{cclcccccccc}
\hline
\multicolumn{2}{c}{}                           &  & \multicolumn{2}{c}{Ekman6} & \multicolumn{2}{c}{VideoEmotion8} & \multicolumn{2}{c}{Music\_video} & \multicolumn{2}{c}{IEMOCAP*} \\ \cmidrule(l){4-11} 
\multicolumn{2}{c}{\hspace{0.65cm}\multirow{-2}{*}{Method}}   &  & ACC         & WA-F1        & ACC         & WA-F1        & ACC          & WA-F1 & ACC          & WA-F1         \\ \hline
\hspace{1.4cm}\multirow{1}{*}{AV-CANet}                               &  &  & 55.63       & 55.44      & 49.53       & 48.24      & 84.81        & 84.63  & 68.73 & 68.73     \\
\hspace{1.4cm}\multirow{2}{*}{AV-CANet$\dagger$} &  &  & 56.56 & 56.12 & 52.34 & 51.06 & 85.32 & 85.20 & 69.43 & 69.27 \\ 
& &  & (\textbf{+0.93}) & (\textbf{+0.68}) & (\textbf{+2.81}) & (\textbf{+2.82}) & (\textbf{+0.51}) & (\textbf{+0.57}) & (\textbf{+0.70}) & (\textbf{+0.54}) \\
\hline
\end{tabular}
}
\end{center}
\vspace{-1.2em}
\end{table*}

\begin{table}[t!]
\centering
\setlength{\arrayrulewidth}{0.22pt}
\caption{Performance comparisons of different initialization weights for AV-CANet on e\_T.}
\vspace{0.05em}
\label{tab_weights}
\small
\renewcommand{\arraystretch}{1}
\resizebox{\linewidth}{!}{%
\begin{tabular}{cccccc}
\hline
EM 
                               & ImageNet \cite{59_deng2009imagenet}          & Ek6 & Ve8 & Mv & IEMOCAP \\ \hline
ACC                            & \textbf{67.00}  & 66.30   & 64.80 & 65.80       & 64.90         \\
WA-F1                          & \textbf{66.28}  & 65.76   & 64.20 & 65.33       & 64.16         \\ \hline
\end{tabular}
}
\vspace{-0.2em}
\end{table}

\begin{figure*}[t!]
\setlength{\belowcaptionskip}{-0.1cm}
\centering
\includegraphics[height=3.45cm, width=\textwidth]{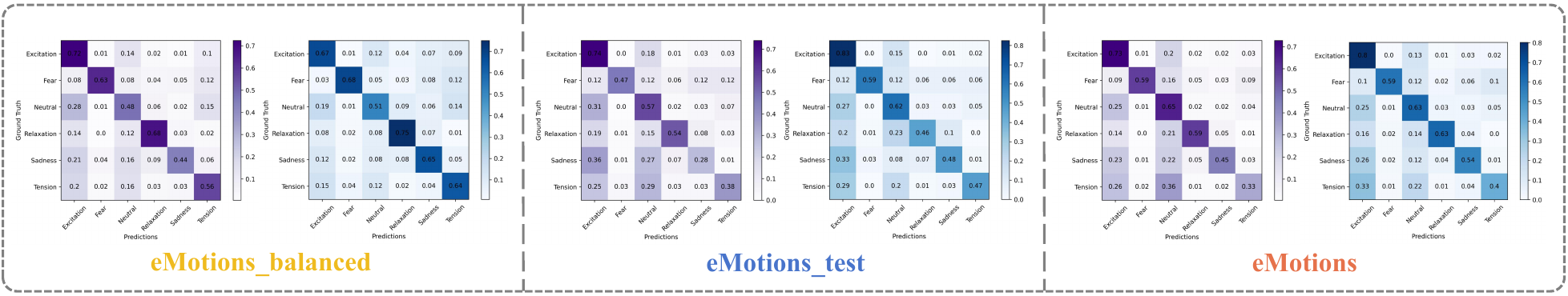}
\vspace{-0.8em}
\caption{The confusion matrices of baseline~\cite{35_zhao2020end} (Purple) and proposed AV-CANet (Blue) on three eMotions-related datasets.}
\label{fig_supp_cm}
\vspace{-0.3em}
\end{figure*}

\begin{figure}[t!]
\setlength{\belowcaptionskip}{-1.5em}
\centering
\includegraphics[scale=0.416]{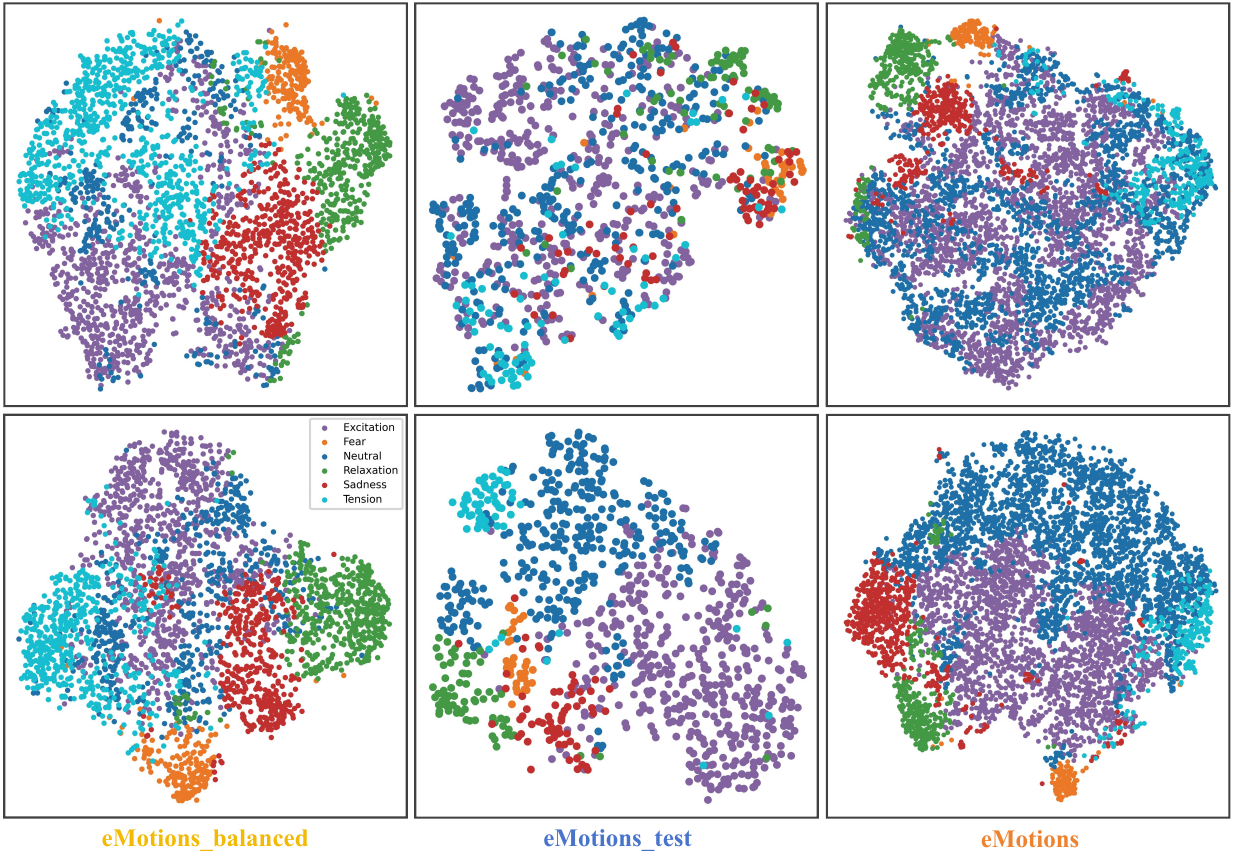}
\caption{The visualizations of embedding spaces on three eMotions-related datasets using t-SNE \cite{new_53_van2008visualizing}. The first and sceond rows respectively are VAANet \cite{35_zhao2020end} and our AV-CANet.}
\label{fig_supp_t-sne}
\end{figure}

\subsection{Cross-datasets Studies}
We conduct cross-datasets experiments using previously learned weights for initializations, and all the parameters are loaded except for those of the final classifier. We first verify the applicability of existing VEA datasets to SVs on e\_T utilizing the weights from four public datasets, as shown in Tab.~\ref{tab_weights}. We observe that when loading the original weights from ImageNet \cite{59_deng2009imagenet}, AV-CANet performs best, suggesting that despite the videos of existing VEA datasets containing more emotional cues, they are inappropriate for the SVs scenario. These outcomes verify the necessity of constructing a dedicated dataset for SVs. 

We then investigate the influence of the more reliable annotations of eMotions on four public datasets using the weights learned from eMotions, as illustrated in Tab.~\ref{cross-dataset}. It can be clearly seen that AV-CANet achieves better performance across all the datasets, exhibiting the largest improvement of 2.81\% ACC and 2.82 WA-F1. This indicates that the more reliably annotated emotions of our dataset can assist models in determining more precise feature mappings and understanding various elements of emotional expressions.

\subsection{Qualitative Experiments}

To further qualitatively show the superiority of our proposed AV-CANet over the baseline model, we compare the confusion matrices and visualize the learned embedding spaces using t-SNE \cite{new_53_van2008visualizing} on three eMotions-related datasets. We can figure out that the prediction accuracies of AV-CANet are generally enhanced when compared to the baseline method~\cite{35_zhao2020end}, as illustrated in Fig.~\ref{fig_supp_cm}. Furthermore, the improvements are particularly significant in some emotional categories, such as the prediction accuracy of Sadness on eMotions\_test increases from 28.00\% to 48.00\%. From Fig.~\ref{fig_supp_t-sne}, we observe that the learned embeddings of our AV-CANet are more discriminative than baseline across three datasets, as evidenced by its more compact intra-class distributions. These observations demonstrate that our AV-CANet, which targetedly tackles the inherent challenges of eMotions, can learn more emotion-related representations for different categories in SVs. Meanwhile, the results also indicate the necessity of designing deep models and strategies towards the category-imbalanced scenarios.

%% file: Sec/5_conclusions.tex
\section{Conclusion and Discussions}
\label{sec:discussion}

In this paper, we propose eMotions, the first large-scale video emotion analysis (VEA) dataset for short-form videos (SVs). It includes 27,996 videos annotated with six emotions, collected from three SV platforms. Meanwhile, we make efforts to augment the labeling quality by alleviating the influence of subjectivities. Additionally, two variant datasets are provided through targeted data sampling. We also develop the baseline method termed AV-CANet to tackle the inherent challenges of VEA towards SVs. Extensive experiments on three eMotions-related datasets and four public datasets verify the superiority of our model. We hope this work can serve as a foundation and inspire more related research.

Looking ahead, we highlight the emerging trends such as audio granularity and text overlay usage in SVs, and aim to address challenges like limited cultural generalizability and the long-tail distributions in the future developments.

%% file: main.bbl
\begin{thebibliography}{10}
\providecommand{\url}[1]{#1}
\csname url@samestyle\endcsname
\providecommand{\newblock}{\relax}
\providecommand{\bibinfo}[2]{#2}
\providecommand{\BIBentrySTDinterwordspacing}{\spaceskip=0pt\relax}
\providecommand{\BIBentryALTinterwordstretchfactor}{4}
\providecommand{\BIBentryALTinterwordspacing}{\spaceskip=\fontdimen2\font plus
\BIBentryALTinterwordstretchfactor\fontdimen3\font minus \fontdimen4\font\relax}
\providecommand{\BIBforeignlanguage}[2]{{%
\expandafter\ifx\csname l@#1\endcsname\relax
\typeout{** WARNING: IEEEtran.bst: No hyphenation pattern has been}%
\typeout{** loaded for the language `#1'. Using the pattern for}%
\typeout{** the default language instead.}%
\else
\language=\csname l@#1\endcsname
\fi
#2}}
\providecommand{\BIBdecl}{\relax}
\BIBdecl

\bibitem{new_21_liu2022ser30k}
S.~Liu, X.~Zhang, and J.~Yang, ``Ser30k: A large-scale dataset for sticker emotion recognition,'' in \emph{Proceedings of the 30th ACM International Conference on Multimedia}, 2022, pp. 33--41.

\bibitem{yang2022disentangled}
D.~Yang, S.~Huang, H.~Kuang, Y.~Du, and L.~Zhang, ``Disentangled representation learning for multimodal emotion recognition,'' in \emph{Proceedings of the 30th ACM international conference on multimedia}, 2022, pp. 1642--1651.

\bibitem{new_54_sobkowicz2012opinion}
P.~Sobkowicz, M.~Kaschesky, and G.~Bouchard, ``Opinion mining in social media: Modeling, simulating, and forecasting political opinions in the web,'' \emph{Government information quarterly}, vol.~29, no.~4, pp. 470--479, 2012.

\bibitem{5_schuller2018age}
D.~Schuller and B.~W. Schuller, ``The age of artificial emotional intelligence,'' \emph{Computer}, vol.~51, no.~9, pp. 38--46, 2018.

\bibitem{depression_recognition}
J.~Ye, Y.~Yu, L.~Lu, H.~Wang, Y.~Zheng, Y.~Liu, and Q.~Wang, ``Dep-former: Multimodal depression recognition based on facial expressions and audio features via emotional changes,'' \emph{IEEE Transactions on Circuits and Systems for Video Technology}, vol.~35, no.~3, pp. 2087--2100, 2025.

\bibitem{new_1_wang2020suppressing}
K.~Wang, X.~Peng, J.~Yang, S.~Lu, and Y.~Qiao, ``Suppressing uncertainties for large-scale facial expression recognition,'' in \emph{Proceedings of the IEEE/CVF conference on computer vision and pattern recognition}, 2020, pp. 6897--6906.

\bibitem{new_3_yang2021circular}
J.~Yang, J.~Li, L.~Li, X.~Wang, and X.~Gao, ``A circular-structured representation for visual emotion distribution learning,'' in \emph{Proceedings of the IEEE/CVF Conference on Computer Vision and Pattern Recognition}, 2021, pp. 4237--4246.

\bibitem{facial_expression_recognition}
Y.~Li, Y.~Gao, B.~Chen, Z.~Zhang, G.~Lu, and D.~Zhang, ``Self-supervised exclusive-inclusive interactive learning for multi-label facial expression recognition in the wild,'' \emph{IEEE Transactions on Circuits and Systems for Video Technology}, vol.~32, no.~5, pp. 3190--3202, 2022.

\bibitem{1_plutchik1994psychology}
R.~Plutchik, \emph{The psychology and biology of emotion.}\hskip 1em plus 0.5em minus 0.4em\relax HarperCollins College Publishers, 1994.

\bibitem{35_zhao2020end}
S.~Zhao, Y.~Ma, Y.~Gu, J.~Yang, T.~Xing, P.~Xu, R.~Hu, H.~Chai, and K.~Keutzer, ``An end-to-end visual-audio attention network for emotion recognition in user-generated videos,'' in \emph{Proceedings of the AAAI Conference on Artificial Intelligence}, vol.~34, no.~01, 2020, pp. 303--311.

\bibitem{tcsvt_CNN-based}
X.~Zhang, F.~Zhang, and C.~Xu, ``Joint expression synthesis and representation learning for facial expression recognition,'' \emph{IEEE Transactions on Circuits and Systems for Video Technology}, vol.~32, no.~3, pp. 1681--1695, 2022.

\bibitem{19_liu2022video}
Z.~Liu, J.~Ning, Y.~Cao, Y.~Wei, Z.~Zhang, S.~Lin, and H.~Hu, ``Video swin transformer,'' in \emph{Proceedings of the IEEE/CVF conference on computer vision and pattern recognition}, 2022, pp. 3202--3211.

\bibitem{ViC-Bench}
X.~Wu, J.~Liu, D.~Huang, X.~Li, Y.~Wang, C.~Chen, L.~Ma, X.~Cao, and J.~Xue, ``Vic-bench: Benchmarking visual-interleaved chain-of-thought capability in mllms with free-style intermediate state representations,'' \emph{arXiv preprint arXiv:2505.14404}, 2025.

\bibitem{new_55_yang2024emogen}
J.~Yang, J.~Feng, and H.~Huang, ``Emogen: Emotional image content generation with text-to-image diffusion models,'' in \emph{Proceedings of the IEEE/CVF Conference on Computer Vision and Pattern Recognition}, 2024, pp. 6358--6368.

\bibitem{AVF-MAE++}
X.~Wu, H.~Sun, Y.~Wang, J.~Nie, J.~Zhang, Y.~Wang, J.~Xue, and L.~He, ``Avf-mae++: Scaling affective video facial masked autoencoders via efficient audio-visual self-supervised learning,'' in \emph{Proceedings of the Computer Vision and Pattern Recognition Conference}, 2025, pp. 9142--9153.

\bibitem{yang2024asynchronous}
D.~Yang, M.~Li, L.~Qu, K.~Yang, P.~Zhai, S.~Wang, and L.~Zhang, ``Asynchronous multimodal video sequence fusion via learning modality-exclusive and-agnostic representations,'' \emph{IEEE Transactions on Circuits and Systems for Video Technology}, 2024.

\bibitem{new_56_lian2023explainable}
Z.~Lian, L.~Sun \emph{et~al.}, ``Explainable multimodal emotion reasoning,'' \emph{arXiv preprint arXiv:2306.15401}, 2023.

\bibitem{wu2025towards}
X.~Wu, H.~Sun, J.~Xue, J.~Nie, X.~Kong, R.~Zhai, D.~Huang, and L.~He, ``Towards emotion analysis in short-form videos: A large-scale dataset and baseline,'' in \emph{Proceedings of the 2025 International Conference on Multimedia Retrieval}, 2025, pp. 1497--1506.

\bibitem{new_9_gunes2006bimodal}
H.~Gunes and M.~Piccardi, ``A bimodal face and body gesture database for automatic analysis of human nonverbal affective behavior,'' in \emph{18th International conference on pattern recognition (ICPR'06)}, vol.~1.\hskip 1em plus 0.5em minus 0.4em\relax IEEE, 2006, pp. 1148--1153.

\bibitem{26_busso2008iemocap}
C.~Busso, M.~Bulut \emph{et~al.}, ``Iemocap: Interactive emotional dyadic motion capture database,'' \emph{Language resources and evaluation}, vol.~42, pp. 335--359, 2008.

\bibitem{new_6_2012Collecting}
A.~Dhall, R.~Goecke, S.~Lucey, and T.~Gedeon, ``Collecting large, richly annotated facial-expression databases from movies,'' \emph{IEEE Multimedia}, vol.~19, no.~3, p. 0034, 2012.

\bibitem{new_8_kollias2018aff}
D.~Kollias and S.~Zafeiriou, ``Aff-wild2: Extending the aff-wild database for affect recognition,'' \emph{arXiv preprint arXiv:1811.07770}, 2018.

\bibitem{12_jiang2014predicting}
Y.-G. Jiang, B.~Xu, and X.~Xue, ``Predicting emotions in user-generated videos,'' in \emph{Proceedings of the AAAI conference on artificial intelligence}, vol.~28, no.~1, 2014.

\bibitem{13_xu2016video}
B.~Xu, Y.~Fu, Y.-G. Jiang, B.~Li, and L.~Sigal, ``Video emotion recognition with transferred deep feature encodings,'' in \emph{proceedings of the 2016 ACM on international conference on multimedia retrieval}, 2016, pp. 15--22.

\bibitem{16_zadeh2018multimodal}
A.~B. Zadeh, P.~P. Liang, S.~Poria, E.~Cambria, and L.-P. Morency, ``Multimodal language analysis in the wild: Cmu-mosei dataset and interpretable dynamic fusion graph,'' in \emph{Proceedings of the 56th Annual Meeting of the Association for Computational Linguistics}, 2018, pp. 2236--2246.

\bibitem{18_lee2019context}
J.~Lee, S.~Kim, S.~Kim, J.~Park, and K.~Sohn, ``Context-aware emotion recognition networks,'' in \emph{Proceedings of the IEEE/CVF international conference on computer vision}, 2019, pp. 10\,143--10\,152.

\bibitem{33_soujanya2018multimodal}
S.~Poria, D.~Hazarika, N.~Majumder, G.~Naik, E.~Cambria, and R.~Mihalcea, ``Meld: A multimodal multi-party dataset for emotion recognition in conversations,'' \emph{arXiv preprint arXiv:1810.02508}, 2018.

\bibitem{7_pandeya2021deep}
Y.~R. Pandeya \emph{et~al.}, ``Deep learning-based late fusion of multimodal information for emotion classification of music video,'' \emph{Multimedia Tools and Applications}, vol.~80, pp. 2887--2905, 2021.

\bibitem{new_58_jiang2020dfew}
X.~Jiang, Y.~Zong, W.~Zheng, C.~Tang, W.~Xia, C.~Lu, and J.~Liu, ``Dfew: A large-scale database for recognizing dynamic facial expressions in the wild,'' in \emph{Proceedings of the 28th ACM international conference on multimedia}, 2020, pp. 2881--2889.

\bibitem{new_57_liu2022mafw}
Y.~Liu, W.~Dai, C.~Feng, W.~Wang, G.~Yin, J.~Zeng, and S.~Shan, ``Mafw: A large-scale, multi-modal, compound affective database for dynamic facial expression recognition in the wild,'' in \emph{Proceedings of the 30th ACM International Conference on Multimedia}, 2022, pp. 24--32.

\bibitem{new_42_lian2024mer}
Z.~Lian, H.~Sun, L.~Sun, Z.~Wen, S.~Zhang, S.~Chen, H.~Gu, J.~Zhao, Z.~Ma, X.~Chen \emph{et~al.}, ``Mer 2024: Semi-supervised learning, noise robustness, and open-vocabulary multimodal emotion recognition,'' \emph{arXiv preprint arXiv:2404.17113}, 2024.

\bibitem{new_34_zhao2021former}
Z.~Zhao and Q.~Liu, ``Former-dfer: Dynamic facial expression recognition transformer,'' in \emph{Proceedings of the 29th ACM International Conference on Multimedia}, 2021, pp. 1553--1561.

\bibitem{new_12_xue2022coarse}
F.~Xue, Z.~Tan, Y.~Zhu, Z.~Ma, and G.~Guo, ``Coarse-to-fine cascaded networks with smooth predicting for video facial expression recognition,'' in \emph{Proceedings of the IEEE/CVF Conference on Computer Vision and Pattern Recognition}, 2022, pp. 2412--2418.

\bibitem{new_36_zhao2023prompting}
Z.~Zhao and I.~Patras, ``Prompting visual-language models for dynamic facial expression recognition,'' \emph{arXiv preprint arXiv:2308.13382}, 2023.

\bibitem{new_43_radford2021learning}
A.~Radford, J.~W. Kim, C.~Hallacy, A.~Ramesh, G.~Goh, S.~Agarwal, G.~Sastry, A.~Askell, P.~Mishkin, J.~Clark \emph{et~al.}, ``Learning transferable visual models from natural language supervision,'' in \emph{International conference on machine learning}.\hskip 1em plus 0.5em minus 0.4em\relax PMLR, 2021, pp. 8748--8763.

\bibitem{40_zhang2023weakly}
Z.~Zhang, L.~Wang, and J.~Yang, ``Weakly supervised video emotion detection and prediction via cross-modal temporal erasing network,'' in \emph{Proceedings of the IEEE/CVF Conference on Computer Vision and Pattern Recognition}, 2023, pp. 18\,888--18\,897.

\bibitem{new_13_mocanu2023multimodal}
B.~Mocanu, R.~Tapu \emph{et~al.}, ``Multimodal emotion recognition using cross modal audio-video fusion with attention and deep metric learning,'' \emph{Image and Vision Computing}, vol. 133, p. 104676, 2023.

\bibitem{new_35_su2020msaf}
L.~Su, C.~Hu \emph{et~al.}, ``Msaf: Multimodal split attention fusion,'' \emph{arXiv preprint arXiv:2012.07175}, 2020.

\bibitem{new_16_tran2022pre}
M.~Tran and M.~Soleymani, ``A pre-trained audio-visual transformer for emotion recognition,'' in \emph{ICASSP 2022-2022 IEEE International Conference on Acoustics, Speech and Signal Processing (ICASSP)}.\hskip 1em plus 0.5em minus 0.4em\relax IEEE, 2022, pp. 4698--4702.

\bibitem{new_14_praveen2023audio}
R.~G. Praveen, P.~Cardinal, and E.~Granger, ``Audio-visual fusion for emotion recognition in the valence-arousal space using joint cross-attention,'' \emph{IEEE Transactions on Biometrics, Behavior, and Identity Science}, 2023.

\bibitem{50_peng2022balanced}
X.~Peng, Y.~Wei, A.~Deng, D.~Wang, and D.~Hu, ``Balanced multimodal learning via on-the-fly gradient modulation,'' in \emph{Proceedings of the IEEE/CVF Conference on Computer Vision and Pattern Recognition}, 2022, pp. 8238--8247.

\bibitem{53_xu2023mmcosine}
R.~Xu, R.~Feng, S.-X. Zhang, and D.~Hu, ``Mmcosine: Multi-modal cosine loss towards balanced audio-visual fine-grained learning,'' in \emph{ICASSP 2023-2023 IEEE International Conference on Acoustics, Speech and Signal Processing (ICASSP)}.\hskip 1em plus 0.5em minus 0.4em\relax IEEE, 2023, pp. 1--5.

\bibitem{new_18_chumachenko2022self}
K.~Chumachenko, A.~Iosifidis, and M.~Gabbouj, ``Self-attention fusion for audiovisual emotion recognition with incomplete data,'' in \emph{2022 26th International Conference on Pattern Recognition (ICPR)}.\hskip 1em plus 0.5em minus 0.4em\relax IEEE, 2022, pp. 2822--2828.

\bibitem{55_poria2017context}
S.~Poria, E.~Cambria \emph{et~al.}, ``Context-dependent sentiment analysis in user-generated videos,'' in \emph{Proceedings of the 55th annual meeting of the association for computational linguistics (volume 1: Long papers)}, 2017, pp. 873--883.

\bibitem{new_59_zhang2024mart}
Z.~Zhang, P.~Zhao, E.~Park, and J.~Yang, ``Mart: Masked affective representation learning via masked temporal distribution distillation,'' in \emph{Proceedings of the IEEE/CVF Conference on Computer Vision and Pattern Recognition}, 2024, pp. 12\,830--12\,840.

\bibitem{new_17_schoneveld2021leveraging}
L.~Schoneveld, A.~Othmani, and H.~Abdelkawy, ``Leveraging recent advances in deep learning for audio-visual emotion recognition,'' \emph{Pattern Recognition Letters}, vol. 146, pp. 1--7, 2021.

\bibitem{2_mayer2002mayer}
J.~D. Mayer, P.~Salovey, and D.~R. Caruso, ``Mayer-salovey-caruso emotional intelligence test (msceit) users manual,'' 2002.

\bibitem{new_29_lavitas2021annotation}
L.~Lavitas, O.~Redfield, A.~Lee, D.~Fletcher, M.~Eck, and S.~Janardhanan, ``Annotation quality framework-accuracy, credibility, and consistency,'' in \emph{NEURIPS 2021 Workshop for Data Centric AI}, 2021.

\bibitem{new_28_abercrombie2023consistency}
G.~Abercrombie \emph{et~al.}, ``Consistency is key: Disentangling label variation in natural language processing with intra-annotator agreement,'' \emph{arXiv preprint arXiv:2301.10684}, 2023.

\bibitem{new_27_hallgren2012computing}
K.~A. Hallgren, ``Computing inter-rater reliability for observational data: an overview and tutorial,'' \emph{Tutorials in quantitative methods for psychology}, vol.~8, no.~1, p.~23, 2012.

\bibitem{new_60_posner2005circumplex}
J.~Posner, J.~A. Russell, and B.~S. Peterson, ``The circumplex model of affect: An integrative approach to affective neuroscience, cognitive development, and psychopathology,'' \emph{Development and psychopathology}, vol.~17, no.~3, pp. 715--734, 2005.

\bibitem{new_23_singh2014sampling}
A.~S. Singh and M.~B. Masuku, ``Sampling techniques \& determination of sample size in applied statistics research: An overview,'' \emph{International Journal of economics, commerce and management}, vol.~2, no.~11, pp. 1--22, 2014.

\bibitem{new_24_ribeiro2011crowdsourcing}
F.~Ribeiro, D.~Florencio, and V.~Nascimento, ``Crowdsourcing subjective image quality evaluation,'' in \emph{2011 18th IEEE International Conference on Image Processing}.\hskip 1em plus 0.5em minus 0.4em\relax IEEE, 2011, pp. 3097--3100.

\bibitem{22_he2016deep}
K.~He, X.~Zhang, S.~Ren, and J.~Sun, ``Deep residual learning for image recognition,'' in \emph{Proceedings of the IEEE conference on computer vision and pattern recognition}, 2016, pp. 770--778.

\bibitem{new_37_yu2022mm}
J.~Yu, Y.~Cheng, R.-W. Zhao, R.~Feng, and Y.~Zhang, ``Mm-pyramid: Multimodal pyramid attentional network for audio-visual event localization and video parsing,'' in \emph{Proceedings of the 30th ACM international conference on multimedia}, 2022, pp. 6241--6249.

\bibitem{57_vaswani2017attention}
A.~Vaswani, N.~Shazeer, N.~Parmar, J.~Uszkoreit, L.~Jones, A.~N. Gomez, {\L}.~Kaiser, and I.~Polosukhin, ``Attention is all you need,'' \emph{Advances in neural information processing systems}, vol.~30, 2017.

\bibitem{27_kong2020panns}
Q.~Kong, Y.~Cao, T.~Iqbal, Y.~Wang, W.~Wang, and M.~D. Plumbley, ``Panns: Large-scale pretrained audio neural networks for audio pattern recognition,'' \emph{IEEE/ACM Transactions on Audio, Speech, and Language Processing}, vol.~28, pp. 2880--2894, 2020.

\bibitem{new_32_desplanques2020ecapa}
B.~Desplanques, J.~Thienpondt, and K.~Demuynck, ``Ecapa-tdnn: Emphasized channel attention, propagation and aggregation in tdnn based speaker verification,'' \emph{arXiv preprint arXiv:2005.07143}, 2020.

\bibitem{34_gao2019res2net}
S.-H. Gao, M.-M. Cheng, K.~Zhao, X.-Y. Zhang, M.-H. Yang, and P.~Torr, ``Res2net: A new multi-scale backbone architecture,'' \emph{IEEE transactions on pattern analysis and machine intelligence}, vol.~43, no.~2, pp. 652--662, 2019.

\bibitem{38_wang2023cam++}
H.~Wang, S.~Zheng, Y.~Chen, L.~Cheng, and Q.~Chen, ``Cam++: A fast and efficient network for speaker verification using context-aware masking,'' \emph{arXiv preprint arXiv:2303.00332}, 2023.

\bibitem{43_carreira2017quo}
J.~Carreira and A.~Zisserman, ``Quo vadis, action recognition? a new model and the kinetics dataset,'' in \emph{proceedings of the IEEE Conference on Computer Vision and Pattern Recognition}, 2017, pp. 6299--6308.

\bibitem{new_33_feichtenhofer2019slowfast}
C.~Feichtenhofer, H.~Fan, J.~Malik, and K.~He, ``Slowfast networks for video recognition,'' in \emph{Proceedings of the IEEE/CVF international conference on computer vision}, 2019, pp. 6202--6211.

\bibitem{30_bertasius2021space}
G.~Bertasius \emph{et~al.}, ``Is space-time attention all you need for video understanding?'' in \emph{ICML}, vol.~2, no.~3, 2021, p.~4.

\bibitem{32_tran2015learning}
D.~Tran, L.~Bourdev, R.~Fergus, L.~Torresani, and M.~Paluri, ``Learning spatiotemporal features with 3d convolutional networks,'' in \emph{Proceedings of the IEEE international conference on computer vision}, 2015, pp. 4489--4497.

\bibitem{21_hara2018can}
K.~Hara, H.~Kataoka, and Y.~Satoh, ``Can spatiotemporal 3d cnns retrace the history of 2d cnns and imagenet?'' in \emph{Proceedings of the IEEE conference on Computer Vision and Pattern Recognition}, 2018, pp. 6546--6555.

\bibitem{46_liu2022convnet}
Z.~Liu, H.~Mao, C.-Y. Wu, C.~Feichtenhofer, T.~Darrell, and S.~Xie, ``A convnet for the 2020s,'' in \emph{Proceedings of the IEEE/CVF conference on computer vision and pattern recognition}, 2022, pp. 11\,976--11\,986.

\bibitem{42_wang2020cspnet}
C.-Y. Wang, H.-Y.~M. Liao, Y.-H. Wu, P.-Y. Chen, J.-W. Hsieh, and I.-H. Yeh, ``Cspnet: A new backbone that can enhance learning capability of cnn,'' in \emph{Proceedings of the IEEE/CVF conference on computer vision and pattern recognition workshops}, 2020, pp. 390--391.

\bibitem{65_loshchilov2017decoupled}
I.~Loshchilov \emph{et~al.}, ``Decoupled weight decay regularization,'' \emph{arXiv preprint arXiv:1711.05101}, 2017.

\bibitem{59_deng2009imagenet}
J.~Deng, W.~Dong, R.~Socher \emph{et~al.}, ``Imagenet: A large-scale hierarchical image database,'' in \emph{2009 IEEE conference on computer vision and pattern recognition}.\hskip 1em plus 0.5em minus 0.4em\relax Ieee, 2009, pp. 248--255.

\bibitem{73_chudasama2022m2fnet}
V.~Chudasama, P.~Kar, A.~Gudmalwar, N.~Shah, P.~Wasnik, and N.~Onoe, ``M2fnet: Multi-modal fusion network for emotion recognition in conversation,'' in \emph{Proceedings of the IEEE/CVF Conference on Computer Vision and Pattern Recognition}, 2022, pp. 4652--4661.

\bibitem{new_45_sun2024hicmae}
L.~Sun, Z.~Lian, B.~Liu, and J.~Tao, ``Hicmae: Hierarchical contrastive masked autoencoder for self-supervised audio-visual emotion recognition,'' \emph{Information Fusion}, vol. 108, p. 102382, 2024.

\bibitem{new_52_sun2023mae}
L.~Sun, Z.~Lian \emph{et~al.}, ``Mae-dfer: Efficient masked autoencoder for self-supervised dynamic facial expression recognition,'' in \emph{Proceedings of the 31st ACM International Conference on Multimedia}, 2023, pp. 6110--6121.

\bibitem{new_61_LLaVA_liu2024visual}
H.~Liu, C.~Li \emph{et~al.}, ``Visual instruction tuning,'' \emph{Advances in neural information processing systems}, vol.~36, 2024.

\bibitem{new_62_qwenvl_bai2023qwen}
J.~Bai, S.~Bai, S.~Yang, S.~Wang, S.~Tan, P.~Wang, J.~Lin, C.~Zhou, and J.~Zhou, ``Qwen-vl: A frontier large vision-language model with versatile abilities,'' \emph{arXiv preprint arXiv:2308.12966}, 2023.

\bibitem{new_63_chen2024internvl}
Z.~Chen, J.~Wu, W.~Wang, W.~Su, G.~Chen, S.~Xing, M.~Zhong, Q.~Zhang, X.~Zhu, L.~Lu \emph{et~al.}, ``Internvl: Scaling up vision foundation models and aligning for generic visual-linguistic tasks,'' in \emph{Proceedings of the IEEE/CVF Conference on Computer Vision and Pattern Recognition}, 2024, pp. 24\,185--24\,198.

\bibitem{new_64_video-llama-zhang2023video}
H.~Zhang, X.~Li, and L.~Bing, ``Video-llama: An instruction-tuned audio-visual language model for video understanding,'' \emph{arXiv preprint arXiv:2306.02858}, 2023.

\bibitem{60_kiela2018efficient}
D.~Kiela, E.~Grave, A.~Joulin, and T.~Mikolov, ``Efficient large-scale multi-modal classification,'' in \emph{Proceedings of the AAAI conference on artificial intelligence}, vol.~32, no.~1, 2018.

\bibitem{new_53_van2008visualizing}
L.~Van~der Maaten and G.~Hinton, ``Visualizing data using t-sne.'' \emph{Journal of machine learning research}, vol.~9, no.~11, 2008.

\end{thebibliography}
